\documentclass[runningheads]{llncs}
\usepackage{graphicx}
\usepackage[table]{xcolor} 

\usepackage{tikz}
\usepackage{comment}
\usepackage{amsmath,amssymb}
\usepackage{color}

\usepackage{float}
\usepackage{booktabs}
\usepackage{multirow}
\usepackage{subfigure}
\usepackage[misc]{ifsym}
\usepackage[accsupp]{axessibility}

\usepackage[pagebackref=true,breaklinks=true,letterpaper=true,colorlinks=true,citecolor=cyan,bookmarks=false]{hyperref}
\usepackage[width=122mm,left=23mm,paperwidth=168mm,height=193mm,top=22.3mm,paperheight=237.6mm]{geometry}

\begin{document}

\pagestyle{headings}
\mainmatter

\title{Multi-Curve Translator for High-Resolution Photorealistic Image Translation}
\titlerunning{Multi-Curve Translator}
\author{Yuda Song \and
Hui Qian \and
Xin Du {\textsuperscript \Letter} }
\authorrunning{Song et al.}

\institute{Zhejiang University, Hangzhou, China\\
\email{\{syd,qianhui,duxin\}@zju.edu.cn}}

\maketitle

\begin{abstract}
    The dominant image-to-image translation methods are based on fully convolutional networks, which extract and translate an image's features and then reconstruct the image.
    However, they have unacceptable computational costs when working with high-resolution images.
    To this end, we present the Multi-Curve Translator (MCT), which not only predicts the translated pixels for the corresponding input pixels but also for their neighboring pixels.
    And if a high-resolution image is downsampled to its low-resolution version, the lost pixels are the remaining pixels' neighboring pixels.
    So MCT makes it possible to feed the network only the downsampled image to perform the mapping for the full-resolution image, which can dramatically lower the computational cost.
    Besides, MCT is a plug-in approach that utilizes existing base models and requires only replacing their output layers.
    Experiments demonstrate that the MCT variants can process 4K images in real-time and achieve comparable or even better performance than the base models on various photorealistic image-to-image translation tasks.
\end{abstract}

\section{Introduction}

Image-to-image (I2I) translation aims to translate images from a source domain to a target domain.
Many computer vision tasks, such as image denoising~\cite{zhang2017beyond}, dehazing~\cite{cai2016dehazenet}, colorization~\cite{zhang2016colorful}, attribute editing~\cite{choi2018stargan}, and style transfer~\cite{gatys2016image}, can be posed as I2I translation problems.
Some approaches~\cite{isola2017image,zhu2017unpaired,liu2017unsupervised,huang2018multimodal,park2020contrastive} use a universal framework to handle various I2I translation problems.
No matter the training scheme and the addressed problem, their network architectures are generally based on fully convolutional networks (FCNs)~\cite{long2015fully}.
However, the computational cost of FCN is proportional to the input image pixels, making high-resolution (HR) images be a considerable obstacle to employing these methods.
For example, CycleGAN~\cite{zhu2017unpaired} requires 56.8G multiply-accumulate operations (MACs) to process a $256 \times 256$ image and requires 7.2T MACs when working with a 4K ($3840 \times 2160$) image, which is unacceptable even for high-performance GPUs.

To this end, some researchers design lightweight networks~\cite{anokhin2020high,liang2021high} or employ model compression~\cite{shu2019co,li2020gan} to save computational cost.
However, designing and training a new lightweight FCN is not easy since it involves a trade-off between efficiency and effectiveness. 
And repeating this procedure for every I2I translation task can be highly time-consuming and power-consuming.
Therefore, we prefer to propose a more flexible approach to the problem.
We found that some photorealistic I2I translation methods~\cite{laffont2014transient,luan2017deep,li2018closed} apply post-processing techniques that constrain the mapping to be spatially smooth to preserve the image's structure information.
So why not just predict a spatially smooth mapping to approximate this translation process in the image space?
We can downsample HR images and use the downsampled images to predict the mappings for the original images.
In this way, we can feed low-resolution (LR) images to the backbone networks, which exponentially reduces the computational cost.
Besides, we can try to reuse the existing FCN architectures.

\begin{figure}[t]
    \centering
    \includegraphics[width=0.8\columnwidth]{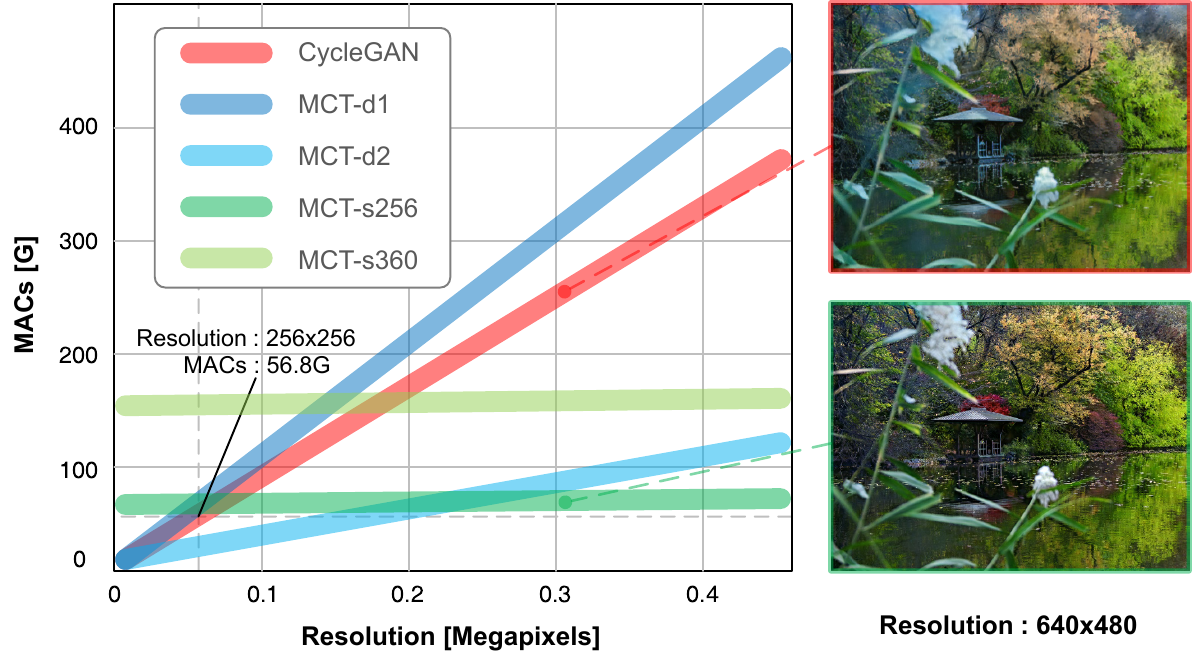}
    \caption{
        An example comparing CycleGAN and its MCT variant on \texttt{autumn2summer}. 
        The only difference between two models is the output layer.
        {\scriptsize\textsf{d2}} means the input image of the backbone network is downsampled by a factor of $2$, and {\scriptsize\textsf{s256}} means the input image of the backbone network is downsampled to $256 \times 256$.
    }
    \label{fig:intro}
\end{figure}

At this point, the key to the problem lies in designing a mapping that can provide sufficient I2I translation capability but has a much lower computational cost than the backbone network.
To address the above challenges, we propose an I2I translator, dubbed \textbf{M}ulti-\textbf{C}urve \textbf{T}ranslator (MCT).
Specifically, we take an existing FCN as the backbone network and find that it only predicts the output pixels for their corresponding input pixels.
So we increase its last layer's output channels to make each output pixel indicate a set of mapping functions in the form of curves.
We quantize these curves as look-up tables (LUTs)~\cite{karaimer2016software,lin2012nonuniform}, then given the output pixel (\emph{i.e.}, curves' parameters) responding to the input pixel of the downsampled image, we can derive the output pixels for all pixels in the full-resolution image's corresponding region.
Besides reducing the computational cost, MCT has additional advantages.
Firstly, an FCN's receptive field is limited, so it may not extract meaningful semantic information when processing HR images.
But for MCT, we can adjust the downsampling ratio to change the backbone network's receptive field size dynamically.
Secondly, since MCT only requires increasing the output channels of the FCN, it is easy to employ it on another I2I translation task without designing a new network architecture.

We extended some I2I translation models to their MCT variants and found that they have significant advantages in saving computational cost and preserving details.
Fig.~\ref{fig:intro} illustrates the performance comparison between CycleGAN and its MCT variant.
Because we can increase the downsampling ratio to reduce the computational cost, the MCT-CycleGAN can always be less computationally intensive than CycleGAN.
In practice, the input images of the MCT's backbone network are downsampled to $256 \times 256$ ({\small\textsf{MCT-s256}} in Fig.~\ref{fig:intro}), consistent with the training set's image size to minimize the gap between inference and training.
In this case, the gap between CycleGAN and MCT-CycleGAN becomes increasingly large as the input image size grows.
Specifically, when processing 4K images, the computational cost of MCT-CycleGAN is only 0.8\% of that of CycleGAN, leading to the former being $40 \times$ faster than the latter on GPUs.
Finally, MCT enables the input image's high-frequency information to flow easily to the output image, making the trees sharper to improve image realism.
While MCT looks appealing, it is to be noted that MCT focuses on photorealistic image-to-image translation and does not work well on more general image-to-image translation tasks, which is our primary future work.

\section{Methodology}

\subsection{Problem Formulation \& Prior Work}

Let $x\in \mathcal{X}$ and $y\in \mathcal{Y}$, the goal of I2I translation is to learn a mapping $G$: $\mathcal{X} \rightarrow \mathcal{Y}$ such that the distribution $p(G(x))$ is as close as possible to the distribution $p(y)$.
Although models for different I2I translation tasks are trained in different manners, they commonly use FCN-based models~\cite{isola2017image,zhu2017unpaired,mechrez2018contextual,park2020contrastive}.
We assume that $G$ is an FCN with weight $\theta$, then the translated image $\tilde{y}$ can be formulated as:
\begin{equation}
    \tilde{y}  = G(x;\theta).
\end{equation}
However, the FCN's computational cost is proportional to the image pixels~\cite{howard2017mobilenets}, and our goal is to break it.
We divide the mapping into two components: the translator $G$ that translates the images from $\mathcal{X}$ to $\mathcal{Y}$ and the encoder $E$ that predicts the translator's parameters.
Assuming that the encoder $E$ with weight $\theta$ encodes the parameters of $G$ from the condition $z$, the translation is:
\begin{equation}
    \tilde{y} = G(x;E(z;\theta)).
\end{equation}

There have been several works based on similar ideas.
Since conventional image processing methods commonly employ filters~\cite{tomasi1998bilateral,he2012guided,yin2019side}, KPN~\cite{mildenhall2018burst} predicts the convolutional filter $G$ with parameter $\theta _i = E(x;\theta)_{i}$ for each pixel $x_i$ and applies it to its spatial support $\Omega(x_i)$ to obtain $\tilde{y}_i$ for burst image denoising.
It can be formulated as:
\begin{equation}
    \tilde{y}_{i} = G(\Omega(x_i);E(x;\theta)_i).
\end{equation}
However, it still requires to perform the FCN on the HR image, leading to no reduction in its computational cost.
Besides, the convolutional filter is still a linear mapping, which has limited translation capability.

Another feasible solution is HyperNetwork~\cite{ha2016hypernetworks}, which uses a network to generate the weights of another network and is originally designed for neural network compression.
Following some HyperNetwork-based works~\cite{fan2018decouple,klocek2019hypernetwork,muller2021overparametrization,shaham2021spatially} on image processing tasks, we can use a encoder $E$ to predict the weights of a lightweight FCN $G$ on the downsampled image $x{\downarrow}$, which can be formulated as:
\begin{equation}
    \tilde{y} = G(x;E(x{\downarrow};\theta)).
\end{equation}
If $E$ is fed with only fixed-size images $x{\downarrow}$, the total computational cost of the model grows slowly with the size of the input images~\cite{song2020model}.
In our experiments, this plain idea works well for photo retouching but not other tasks.

We try to combine the two approaches above to overcome their respective shortcomings.
We expect the encoder $E$ to encode the parameter maps $F = E(x{\downarrow};\theta)$, in which each cell $F_j$ contains a set of translator's parameters:
\begin{equation}
    \tilde{y}_i = G(x_i;F_j).
    \label{eq:mapping}
\end{equation}
Since the parameter maps $F$ cannot be aligned with $x$, we also need to define the relation between $i$ and $j$.
We will detail our MCT in the next subsection.

Similar works to MCT consist of bilateral learning~\cite{gharbi2017deep,xia2020joint,zheng2021ultra}, curve mapping~\cite{kim2020global,li2020flexible,moran2021curl} and 3D LUTs~\cite{zeng2020learning,yang2022adaint}, as they are all based on slicing operation.
In contrast to bilateral learning-based methods, MCT predicts pixel values rather than affine transformations, which allows us to directly constrain the output of MCT to prevent falling into poor solutions when training on unpaired datasets.
Curve-based methods usually predict a global transformation, which prevents them from working on more challenging I2I tasks such as daytime translation.
3D LUTs predict global transformations like curve mappings, but they have a stronger translation capability.
Ideally, we could introduce spatial coordinates to extend 3D LUTs to 5D LUTs, but this would lead to an unacceptable computational cost and memory consumption.
From the implementation perspective, MCT extends the curve-based methods by introducing spatial coordinates and channel interactions to improve translation capability, which can be implemented using 3D LUTs.
More importantly, MCT is a plug-in module that does not rely on fancy backbone networks and loss functions, and it can be trained directly on small-sized images, dramatically reducing the effort to modify the methods.

\subsection{Multi-Curve Translator}

\begin{figure}[t]
    \centering
    \includegraphics[width=1.0\textwidth]{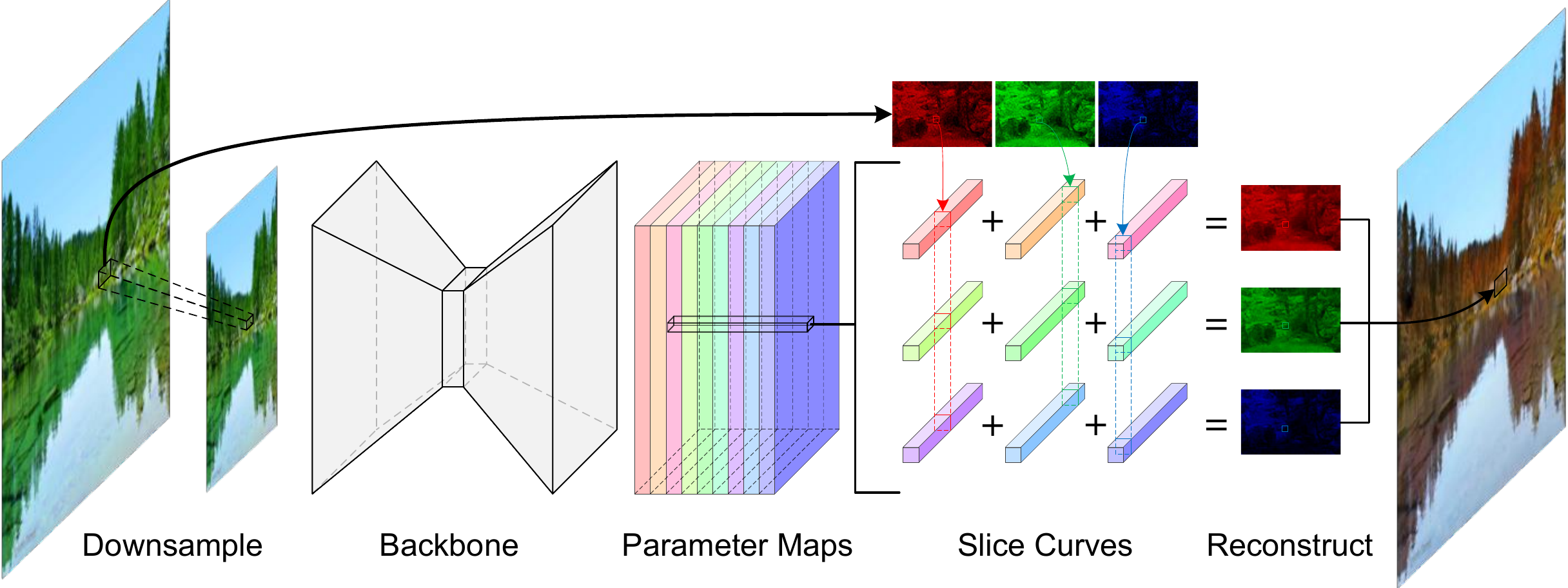}
    \caption{
        Inference workflow of MCT. 
        The backbone network receives the downsampled image and predicts the curve parameter maps with the same spatial size as the downsampled image.  
        A cell of curve parameter maps consists of 9 sets of curves in the form of 1D LUTs, responsible for translating the corresponding region in the HR image. 
    }
    \label{fig:mct}
\end{figure}

Recalling Eq.(\ref{eq:mapping}), our goals are to 1) design the encoder $E$; 2) design the translator $G$; and 3) define the relation between $i$ and $j$.
We desire our approach to be plug-in for the existing I2I translation models.
Since the models of I2I translation tasks are often based on FCNs, we directly use these networks as our base models (\emph{i.e.}, backbone networks) to eliminate the effort involved in designing the encoder $E$.
Then the only modification needed is to increase their last layer's output channels to match the parameters of the translator $G$.
Given that $x \in \mathcal{R}^{H \times W \times 3}$ and $x{\downarrow} \in \mathcal{R}^{H_d \times W_d \times 3}$, then $F \in \mathcal{R}^{H_d \times W_d \times C}$, where $C$ is the number of parameters of $G$.

Reviewing the idea of HyperNetwork, a simple idea is to employ an FCN as the translator $G$.
However, since convolutional layers lead to a large $C$, we seek another expressive nonlinear mapping with fewer parameters to replace the FCN.
We found that some curve-based methods~\cite{kim2020global,li2020flexible,moran2021curl} achieved better performance than FCN-based methods~\cite{chen2018deep,ignatov2017dslr,wei2018deep} on photo retouching.
These curve-based methods make the network regress the knot points of the curve to mimic the color adjustment curve tool.
Although these methods implement knot points in different ways, they are all equivalent to 1D LUTs~\cite{karaimer2016software,lin2012nonuniform}.
We illustrate the transformation function using the curve in the form of a 1D LUT for a grayscale image.
Given a 1D LUT $\mathbf{T} = \{ t_{(k)}\} _{k=0,...,M-1}$ (\emph{i.e.}, $M$ knot points), pixel $x_{(\rm{i,j})}$ can find its location ${\rm{z}}$ in the LUT via a lookup operation:
\begin{equation}
    {\rm{z}} = x_{(\rm{i,j})} \cdot {\textstyle\frac{M - 1}{C_{max}}},
\label{eq:lookup}
\end{equation}
where $C_{max}$ is the maximum pixel value.
Since ${\rm{z}}$ may not be an integer, we should derive the output pixel value via interpolation.
Let $d_{\rm{z}} = {\rm{z}} - \lfloor {\rm{z}}\rfloor$, where $\lfloor\cdot\rfloor$ is the floor function. 
Given that $\lceil\cdot\rceil$ is the ceil function, we derive output pixel value $y_{(\rm{i,j})}$ via linear interpolation:
\begin{equation}
    \tilde{y}_{(\rm{i,j})} = (1-d_{\rm{z}}) \cdot t_{(\lfloor {\rm{z}} \rfloor)} + d_{\rm{z}}  \cdot t_{(\lceil {\rm{z}} \rceil)}.
\label{eq:interpolation}
\end{equation}

Finally we need to define the relation between $i$ and $j$.
We can upsample the parameter maps to make their resolution the same as $x$ (\emph{i.e.}, $F\uparrow \in \mathcal{R}^{H \times W \times C}$).
Unfortunately, while the computational cost of this operation is acceptable, it produces larger parameter maps, which may consume a lot of memory.
Inspired by bilateral grid~\cite{chen2007real}, we employ a 3D LUT $\mathbf{T} \in \mathcal{R}^{H_d \times W_d \times M}$.
Given a grayscale pixel $x_{(\rm{i,j})}$, its location $({\rm x,y,z})$ in the 3D LUT lattice is:
\begin{equation}
\label{eq:lookup2}
    {\rm{x}} = {\rm i} \cdot {\textstyle\frac{H_d - 1}{H - 1}}, \, {\rm{y}} = {\rm j} \cdot {\textstyle\frac{W_d - 1}{W - 1}}, \, {\rm{z}} = x_{(\rm{i,j})} \cdot {\textstyle\frac{M - 1}{C_{max}}}.
\end{equation}
Let $d_{\rm{x}} = {\rm{x}} - \lfloor {\rm{x}}\rfloor$ and $d_{\rm{y}} = {\rm{y}} - \lfloor {\rm{y}}\rfloor$, we extend Eq.(\ref{eq:interpolation}) to trilinear interpolation to slice the output pixel:
\begin{equation}
\begin{split}
\label{eq:trilinear}
\scriptstyle
\tilde{y}_{(\rm{i,j})}
&\scriptstyle= (1-d_{\rm{x}} )(1-d_{\rm{y}} )(1-d_{\rm{z}} ) t_{(\lfloor \! {\rm{x}} \! \rfloor,\lfloor \!{\rm{y}} \! \rfloor,\lfloor \!{\rm{z}} \! \rfloor)} + d_{\rm{x}} d_{\rm{y}} d_{\rm{z}} t_{(\lceil \!  {\rm{x}} \! \rceil,\lceil \!  {\rm{y}} \! \rceil,\lceil \!  {\rm{z}} \! \rceil)} \\
&\scriptstyle+ (1-d_{\rm{x}} )d_{\rm{y}} (1-d_{\rm{z}} )t_{(\lfloor \! {\rm{x}} \! \rfloor,\lceil \!  {\rm{y}} \! \rceil,\lfloor \!{\rm{z}} \! \rfloor)} + d_{\rm{x}} (1-d_{\rm{y}} )d_{\rm{z}} t_{(\lceil \!  {\rm{x}} \! \rceil,\lfloor \!{\rm{y}} \! \rfloor,\lceil \!  {\rm{z}} \! \rceil)}\\
&\scriptstyle + d_{\rm{x}} d_{\rm{y}} (1-d_{\rm{z}} )t_{(\lceil \!  {\rm{x}} \! \rceil,\lceil \!  {\rm{y}} \! \rceil,\lfloor \!{\rm{z}} \! \rfloor)} + (1-d_{\rm{x}} )(1-d_{\rm{y}} )d_{\rm{z}} t_{(\lfloor \! {\rm{x}} \! \rfloor,\lfloor \!{\rm{y}} \! \rfloor,\lceil \!  {\rm{z}} \! \rceil)}  \\
&\scriptstyle + d_{\rm{x}} (1-d_{\rm{y}} )(1-d_{\rm{z}} )t_{(\lceil \!  {\rm{x}} \! \rceil,\lfloor \!{\rm{y}} \! \rfloor,\lfloor \!{\rm{z}} \! \rfloor)} + (1-d_{\rm{x}} )d_{\rm{y}} d_{\rm{z}} t_{(\lfloor \! {\rm{x}} \! \rfloor,\lceil \!  {\rm{y}} \! \rceil,\lceil \!  {\rm{z}} \! \rceil)}.
\end{split}
\end{equation}

For RGB color images, we employ the channel-crossing strategy~\cite{song2021starenhancer}.
Specifically, 9 curves should be learned, corresponding to $\{\mathbf{T}^{p \rightarrow q}\} _{p,q \in \{ R,G,B \}} $ respectively ($C=9M$).
Let $\mathbf{T}(\cdot)$ denote Eq.(\ref{eq:lookup2}-\ref{eq:trilinear}), we derive output pixel via:
\begin{equation}
    \tilde{y}_{(\rm{i,j})}^q = \mathbf{T}^{R \rightarrow q}(x_{(\rm{i,j})}^R) + \mathbf{T}^{G \rightarrow q}(x_{(\rm{i,j})}^G) + \mathbf{T}^{B \rightarrow q}(x_{(\rm{i,j})}^B).
\end{equation}
Since the translator consists of a large number of curves, we call it Multi-Curve Translator (MCT).
Fig.~\ref{fig:mct} shows how MCT processes a HR image.

\subsection{Training Strategy}\label{sec:training}

\begin{figure} [t]
\centering
\subfigure[Base output constraint]{
\includegraphics[width=0.45\textwidth]{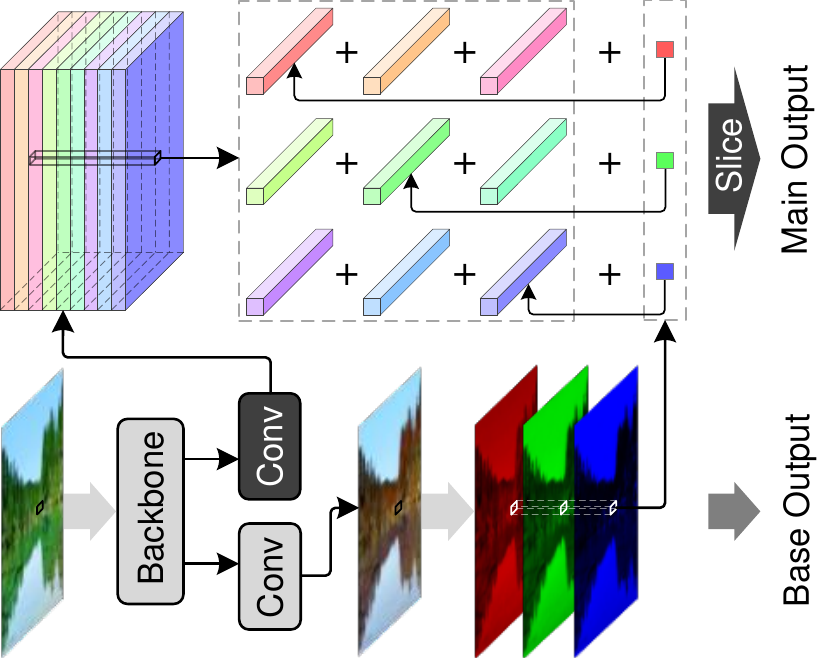}
}
\subfigure[Pixel non-alignment]{
\includegraphics[width=0.45\textwidth]{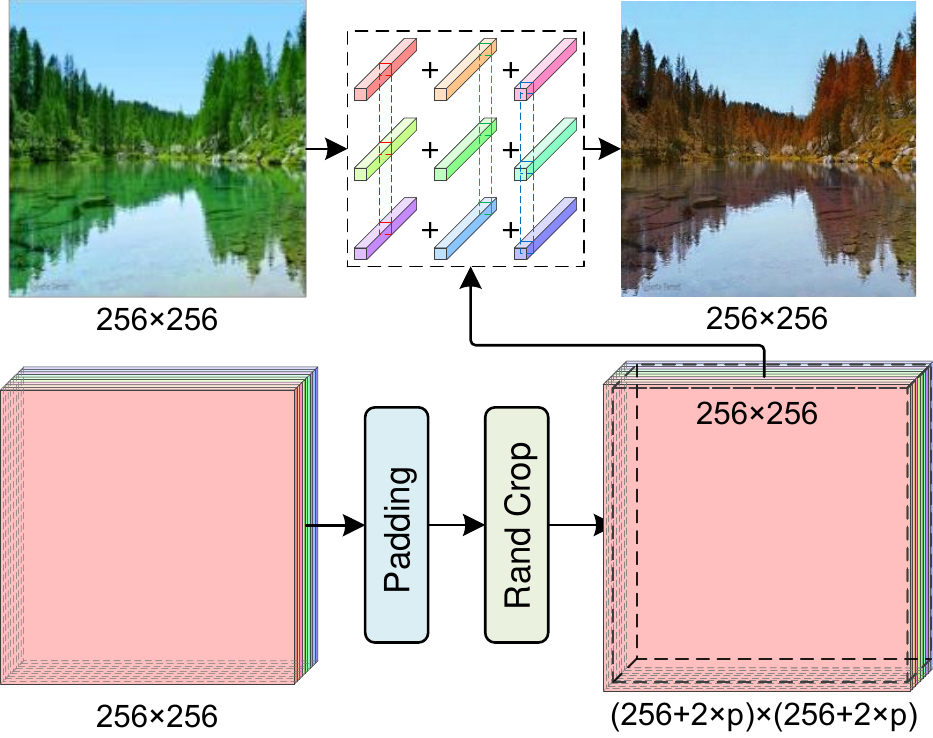}
}
\caption{ \label{fig:train}
Two strategies to train MCT variants in a stable manner.
(a) Use the translated LR image as the base of the parameter maps. 
We can constrain the base output in the training phase to ensure that the backbone network does not degrade.
(b) Train the MCT using only LR images. 
We use padding and random cropping to obtain parameter maps that are not pixel-aligned with the LR images.
}
\end{figure}

Although MCT appears to be more complex than reconstructing images using the convolutional layer, its last layer still outputs pixel values.
As a special case, when only the LUTs $\{\mathbf{T}^{p \rightarrow p}\} _{p \in \{ R,G,B \}} $  are included and $M=1$, MCT is equivalent to upsampling the output image of the base model.
However, we found that the MCT variant is more like to fall into poor solutions than the base model.
We review the MCT and find that the input image's information flows into the output image through the backbone network and slicing operation.
We suppose that the cause of the performance degradation is that the network has difficulty balancing the information flowing through the two routes.
Therefore, we add constraints to the MCT in the training phase to drive the information from both routes to flow adequately into the output image.

Firstly, we should make the MCT leverage the information of downsampled images.
MCT makes high-frequency information from the input image be easily retained in the output image, but we found that MCT may learn a simple color transformation.
The problem arises because the information flows too easily from the input to the output, leading to a ``short circuit'' phenomenon that traps the network in a poor local optimum solution.
Recalling the special case of MCT, we find that $\{\mathbf{T}^{p \rightarrow p}\} _{p \in \{ R,G,B \}} $ can be decomposed into LUTs and biases, as shown in Fig.~\ref{fig:train}(a).
Specifically, we use the last layer of the base model to predict the reconstructed image and add its pixel values as biases to the corresponding LUTs.
So we can obtain the base output $\tilde{y}_{b}$ and the main output $\tilde{y}_{m}$ by $[\tilde{y}_{b}, \tilde{y}_{m}] = [G_b(x), G_m(x)]$ and constrain $\tilde{y}_{b}$ at training phase to ensure that the backbone network does not degrade.
This strategy has a bonus that the pre-trained base model's weight can be fully utilized, including the output layer.
Therefore, we employ this strategy even if we do not need to constrain the base output.
In extreme cases, we can fine-tune only the added output convolutional layers to make the training faster.

Secondly, we should make the MCT leverage the information of HR images.
MCT allows us to perform the backbone network on LR images, dramatically reducing the computational cost of translating HR images.
However, we are still unable to use HR images as training data during training directly.
The reasons are threefold:
1) loading and preprocessing HR images takes a lot of time, resulting in inefficient training;
2) The discriminator's computational cost remains proportional to the input image pixels.
3) Existing datasets often provide low-resolution images.
For this reason, we still use LR images to train the MCT.
As shown in Fig.~\ref{fig:train}(b), we first pad each side of the parameter maps by size $p$ with duplication and randomly crop them to the size before padding, then the image and the parameter maps are not pixel-wise aligned, forcing MCT to extract high-frequency information from the image.
We can also achieve more complex pixel misalignment by adding small random noise to $\rm{x}$ and $\rm{y}$, but this does not visibly improve performance in our experiments.

\section{Applications}

We apply MCT to extend some representative I2I translation methods.
Unless otherwise noted, we set $H_d = W_d = 256$ and $M=8$, and employ pixel unaligned training strategy with $p=1$ but do not constrain the base output.

\subsection{Photorealistic I2I Translation}

We refer here to the I2I translation tasks done with GANs~\cite{goodfellow2014generative}.
We perform the daytime translation (\texttt{day2dusk}) and season translation (\texttt{summer2autumn}) for experiments.
To extend to HR scenes, we collected new unpaired datasets from Flickr\footnote{\url{https://www.flickr.com/}} with image resolutions ranging from 480p to 8K.
Each domain of the datasets contains 2200 images, of which 2000 LR images are downsampled for training and the remaining 200 HR images for testing.

We employ CycleGAN~\cite{zhu2017unpaired} and UNIT~\cite{liu2017unsupervised} as the base models, which use different training procedures.
When training the MCT variants, we add constraints to the base output $y_b$ when updating the generator.
Let the conventional generator's loss function be $\mathcal{L}_{base}$, then the loss function of the MCT variant is $\mathcal{L}=\mathcal{L}_{base} + \lambda \mathcal{L}_{reg}$, where $\lambda=1$ for CycleGAN and $\lambda=10$ for UNIT.
$\mathcal{L}_{reg}$ is a cycle-consistency loss~\cite{zhu2017unpaired,kim2017learning} constrainting the base output:
\begin{equation}
    \mathcal{L}_{reg} = \left\lVert G_b^{y \rightarrow x}(G_m^{x \rightarrow y}(x)) - x \right\rVert _1.
\end{equation}

\subsection{Style Transfer}

Style transfer aims at transferring the style from a reference image to a content image and is divided into two types: artistic style transfer and photorealistic style transfer.
We only study the photorealistic style transfer since it fits our motivation.
We use the Microsoft COCO dataset~\cite{lin2014microsoft} to train the base models and their MCT variants, and the test set consists of the examples provided by DPST~\cite{luan2017deep} with image resolutions ranging from 720p to 4K.

We use AdaIN~\cite{huang2017arbitrary} and WCT$^2$~\cite{yoo2019photorealistic} as the base models since they employ different training schemes.
AdaIN is designed for artistic style transfer, with few constraints on preserving high-frequency information.
It employs a weighted combination of the content loss $\mathcal{L}_c$ and the style loss $\mathcal{L}_s$ with the weight $\lambda$, \emph{i.e.} $\mathcal{L} = \mathcal{L}_c + \lambda \mathcal{L}_s$.
Both $\mathcal{L}_c$ and $\mathcal{L}_s$ use pre-trained VGG-19~\cite{simonyan2014very} to compute the loss function without constraining the pixels of the images.
So we add a gradient loss $\mathcal{L}_g$ to prompt the preservation of the geometric structure:
\begin{equation}
    \mathcal{L}_g = \left\lVert \nabla_h G(x) - \nabla_h x \right\rVert _2^2 + \left\lVert \nabla_v G(x) - \nabla_v x \right\rVert _2^2,
\end{equation}
where $\nabla_h$ ($\nabla_v$) denotes the gradient operator along the horizontal (vertical) direction.
The modified AdaIN's full objective is $\mathcal{L} = \mathcal{L}_c + \lambda_1 \mathcal{L}_s + \lambda_2 \mathcal{L}_g$, where $\lambda_1 = 1$ and $\lambda_2 = 100$.
The WCT$^2$'s scheme is special because it only requires the output image to reconstruct the input during training and performs WCT~\cite{li2017universal} sequentially at each scale to achieve stylization during testing.
Let the reconstruction loss function of WCT$^2$ be $\mathcal{L}_{rec}(G(x), x)$, and its MCT variant's loss function is:
\begin{equation}
    \mathcal{L} = \mathcal{L}_{rec}(G_b(x), x) + \mathcal{L}_{rec}(G_m(x), x).
\end{equation}
WCT$^2$'s training scheme makes the HR image's low-frequency information flow easily to the output image, so we perform the grayscale operation on the HR image.
When performing stylization, we further match the HR image's brightness with the reference image's brightness to prevent the brightness of the HR image from being retained in the output image.

\subsection{Image Dehazing}

Image dehazing aims to recover clean images from hazy images, which is essential for subsequent high-level tasks.
We choose 6000 synthetic image pairs from RESIDE dataset~\cite{li2018benchmarking} for training, 3000 from the ITS subset, and 3000 from the OTS subset.
We use two datasets, named SOTS~\cite{li2018benchmarking} and HazeRD~\cite{zhang2017hazerd}, to evaluate the performance of the methods.
Note that the SOTS has more image pairs while the HazeRD has 4K-resolution image pairs.

We take GCANet~\cite{chen2019gated} and MSBDN~\cite{dong2020multi} as the base models.
GCANet expands the receptive field by dilated convolution~\cite{liang2015semantic}, while MSBDN uses upsampling and downsampling operations.
For simplicity, we use only the $\mathcal{L}_1$ loss function to train the models instead of using the original training scheme of the base models.
Note that for supervised training, the base output constraint is optional.

\subsection{Photo Retouching}

Photo retouching aims to adjust an image's brightness, contrast, and so on to make the image fit people's aesthetics.
We choose 4500 image pairs from the MIT-Adobe-5K dataset~\cite{bychkovsky2011learning} for training and the remaining 500 image pairs for testing with image resolutions ranging from 2K to 6K.
We also employ an unpaired training scheme, with the 2250 images as domain $\mathcal{X}$ and the remaining 2250 images as domain $\mathcal{Y}$ in the training set.

We use DPED~\cite{ignatov2017dslr} and DPE~\cite{chen2018deep} as the base models.
Specifically, DPED uses a residual network~\cite{he2016deep} without downsampling and upsampling, leading to a small receptive field.
But large scale context is critical for photo retouching, which is used to sense the illumination and contrast of an image~\cite{gharbi2017deep,chen2018deep}.
So we set $H_d = W_d = 32$ for DPED to enlarge the receptive field without modifying the network architecture.
Since color mapping is critical for photo retouching, we set $p=8$ for DPE.
For paired training, we use the $\mathcal{L}_1$ loss function to train the models.
We employ the CycleGAN's training scheme for unpaired training rather than the base models' training scheme for comparison purposes.

\section{Experiments}

\subsection{Runtime}

\begin{figure} [t]
    \centering
    \hspace{-0.04\textwidth}
    \subfigure[GCANet]{
    \includegraphics[width=0.5\textwidth]{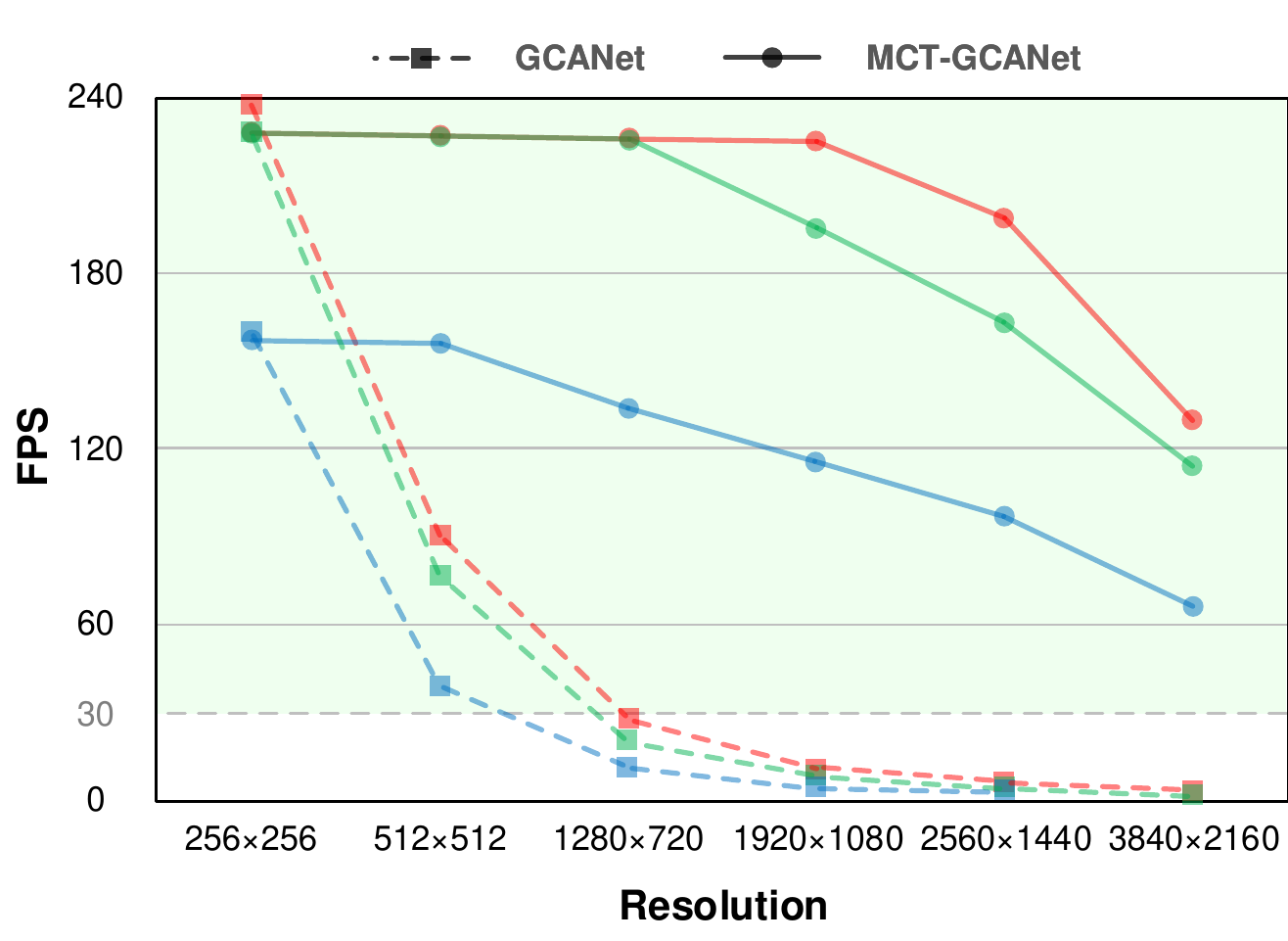}
    }
    \hspace{-0.02\textwidth}
    \subfigure[CycleGAN]{
    \includegraphics[width=0.5\textwidth]{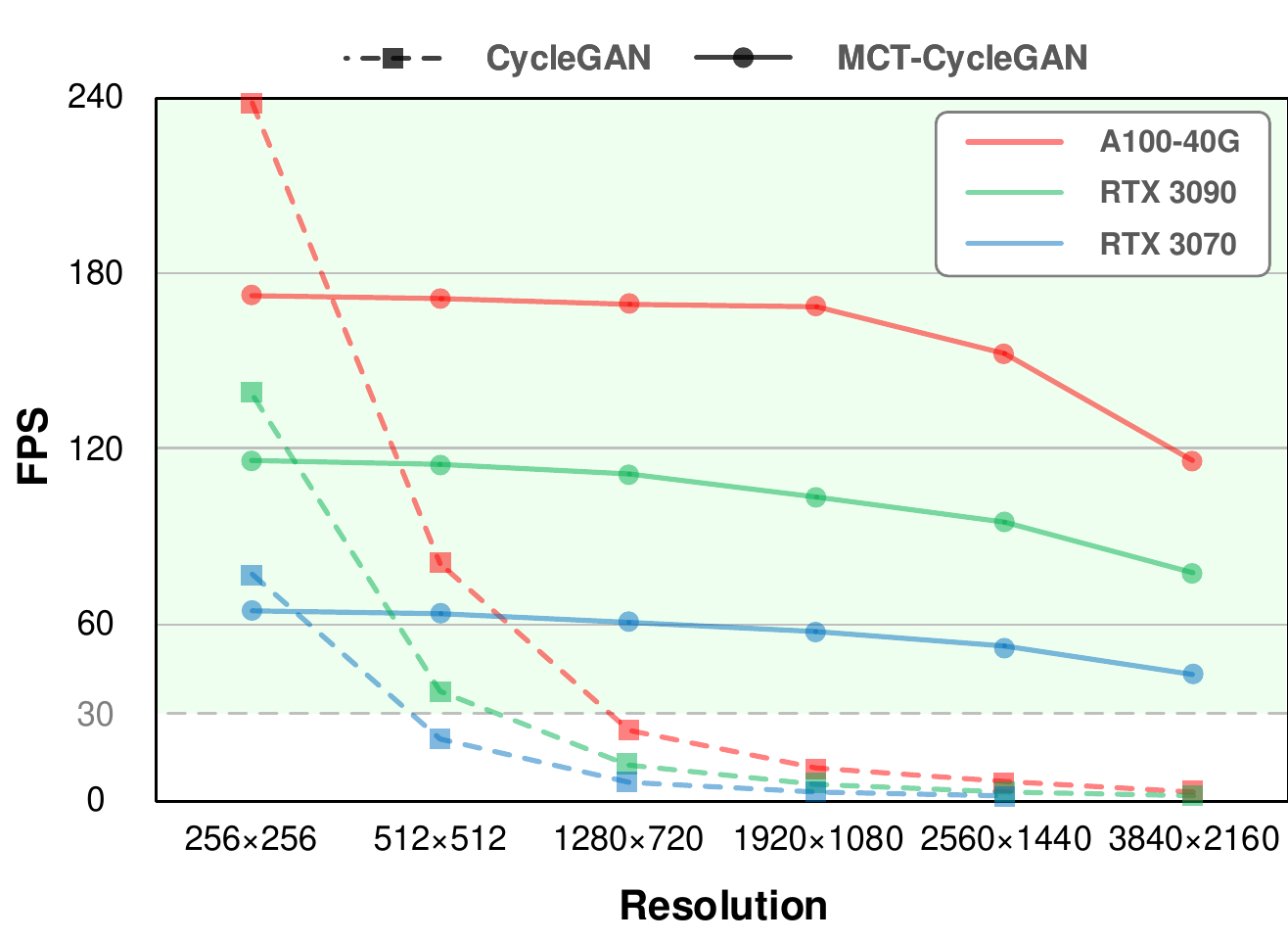}
    }
    \hspace{-0.04\textwidth}
    \caption{ \label{fig:fps}
    Runtime comparison of the base models and their MCT variants on GPUs.
    No data for base models means that they runs with an OOM at that resolution.
}
\end{figure}

We have shown the advantages of the MCT in terms of computational cost in Fig.~\ref{fig:intro}, but MACs are indirect metrics of speed~\cite{ma2018shufflenet}, which is an unconvincing indicator. 
So we test the runtime of the base models and their MCT variants on multiple hardware platforms.
Specifically, we use the PyTorch framework to test each method's frames per second (FPS) in {\small\textsf{float32}} data format and set the mini-batch size to 1.
Given the size of the input image, we randomly generate $200$ images and compute the FPS for a single experiment by recording the total time to process the $200$ images.
Then we repeated each experiment $10$ times and took the median of the $10$ results as the final result.
Fig.~\ref{fig:fps} illustrates the FPS of base models and their variants running on $3$ models of GPUs.

If we feed a $256 \times 256$ image to the models, the MCT variants do not have any advantage over the base models since we set $H_d=W_d=256$ in the experiment.
On the other hand, the large $7 \times 7$ convolution kernels for the output layer introduce an additional 25\% computational cost to the backbone network since the output channels of the MCT-CycleGAN's output layer increase.
Finally, the curve slicing operation contains some operations with low computational cost but high memory access cost (\emph{e.g.} indexing), further increasing the MCT variants' runtime.
Fortunately, the computational cost of MCT's backbone network does not vary with the image size, which gives it a distinct advantage when working with HR images.
When the input image size is $512 \times 512$, the MCT variants are significantly faster than the base models. 
Moreover, the gap between the MCT variants and the base models becomes increasingly large as the input image size grows.
Taking 30 FPS as the cut-off for whether a model can run in real-time, the MCT variants can process 4K images in real-time on $3$ models of GPUs.
As a comparison, CycleGAN takes $40 \times$ longer to process a 4K image (116.0 FPS vs. 2.7 FPS on A100), even with an out-of-memory (OOM) on RTX 3070.
The computational cost of the curve slicing operation is so low that it accounts for less than 1\% of the overall computational cost for processing 4K images. 
Still, it introduces a high memory access cost, making the curve slicing operation limited by the GPU's memory bandwidth.
Finally, MCT-GCANet processes $256 \times 256$ and $512 \times 512$ images at almost the same speed on the A100 and RTX 3090, probably due to the limitations of CPU performance and PyTorch runtime.

\subsection{Qualitative Comparison}

\begin{figure}[htp]
    \centering
    \includegraphics[width=1.0\textwidth]{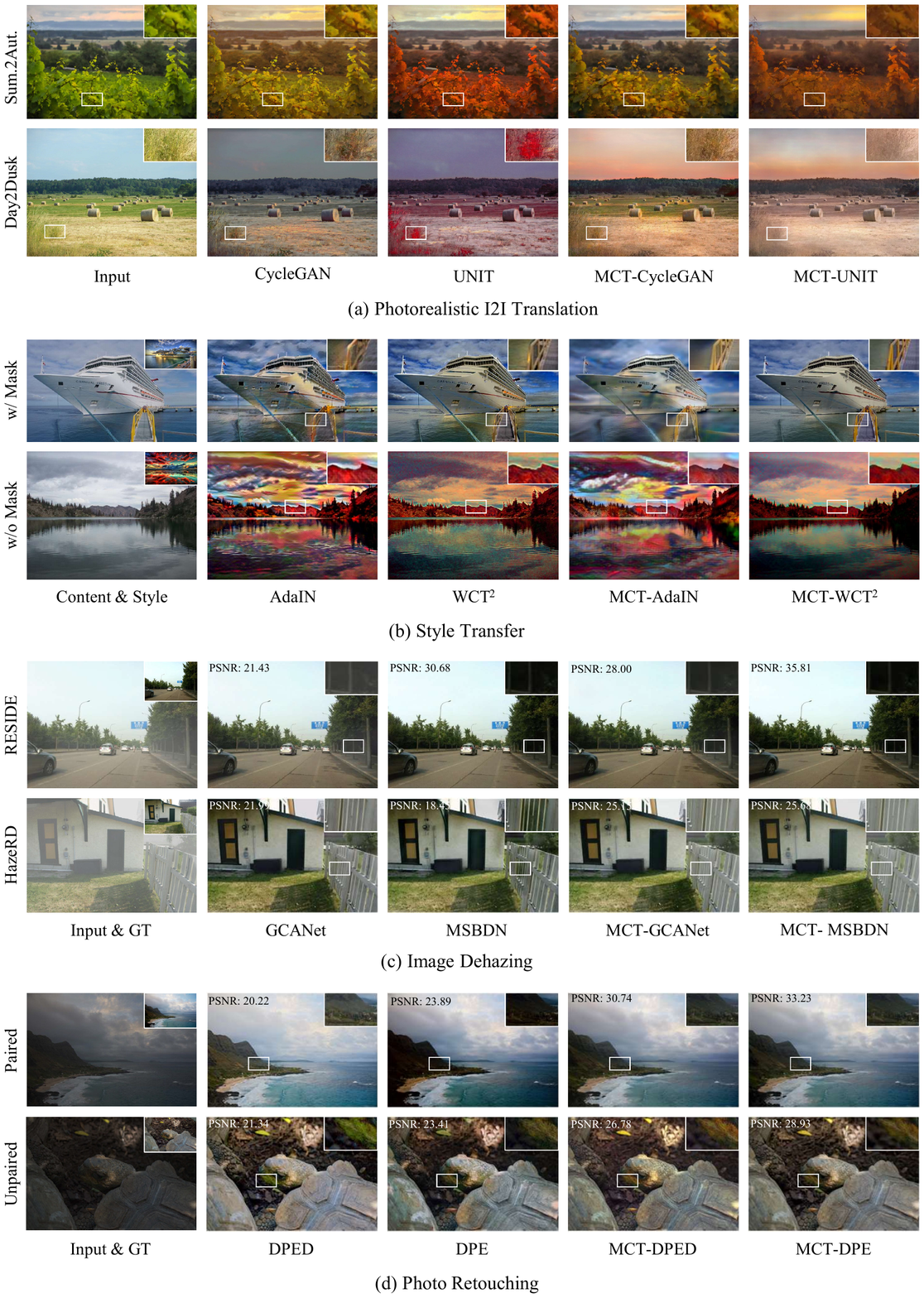}
    \caption{
        Qualitative comparison of four I2I translation tasks, each consisting of examples in two experimental settings.
        From top to bottom are (a) photorealistic I2I translation, (b) style transfer, (c) image dehazing, and (d) photo retouching. 
    }
    \label{fig:exp}
\end{figure}

We qualitatively compare the base models with their MCT variants on four I2I translation tasks, and Fig.~\ref{fig:exp} shows some examples.

\begin{figure*}[t]
    \centering
    \includegraphics[width=1.0\textwidth]{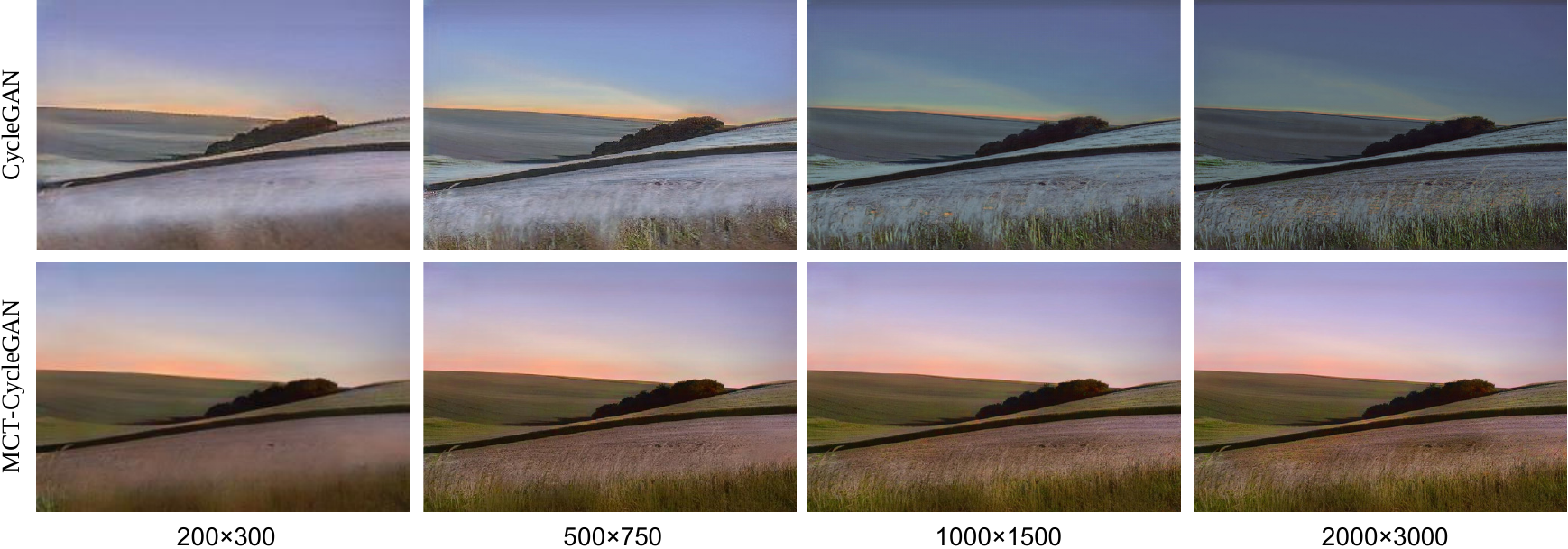}
    \caption{
        Comparison of CycleGAN and MCT-CycleGAN when processing the same image in different resolution versions on \texttt{day2dusk}.
    }
    \label{fig:size}
\end{figure*}

Almost all base models have a non-global and fixed-size receptive field. 
In contrast, the receptive field of MCT variants grows larger as the input image becomes larger.
In the \texttt{day2dusk} example, the base models only change the colors of the ground and sky and do not generate sunset light.
This is because the resolution of the input image is so high that the base models with limited receptive fields cannot determine the sky area.
Fig.~\ref{fig:size} futher shows the results when CycleGAN and MCT-CycleGAN process the same image in different resolution versions.
CycleGAN works fine on low-resolution images but cannot translate HR images well.
The same problem occurs when MSBDN processes HR images in the HazeRD dataset, where significant black artifacts appear on the white railings because the receptive field of MSBDN is not large enough to capture the railings' semantic information.
The DPED's receptive field is small, so it tends to adjust the image's brightness locally to normal brightness, but the image lacks contrast globally.
By lowering the resolution of the downsampled images, the MCT variant of DPED has a large receptive field so that it can better capture global illumination, leading to visually more pleasing results.
In short, the MCT's receptive field is dynamic, which helps to capture the HR images' semantics.

MCT's curve slicing operation allows the backbone network to focus more on region semantics than retaining the high-frequency information.
This is evident in the comparison of AdaIN with its MCT variant.
The original network architecture of AdaIN does not contain any skip connection, resulting in the high-frequency information that VGG-19 loses not being recovered. 
Therefore, even after adjusting the weights of content loss and style loss and introducing the gradient loss, AdaIN still cannot reconstruct the input image's high-frequency information. 
For example, the text and railing are blurred, and the texture of the mountain is lost.
In contrast, its MCT variant can preserve the high-frequency information in the input image by curve slicing operation without being limited by the network architecture.
However, for network architectures like GCANet, the high-frequency information flow is only shifted from skip connections to the curve slicing operation, which does not produce visible differences in the output image details.
We consider that MCT is more likely to retain high-frequency information of the input image.

\subsection{Quantitative Comparison}

\begin{table}[t]
    \renewcommand\arraystretch{1.5}
    \footnotesize
	\centering
	\caption{
        Quantitative comparison (PSNR \& SSIM) of the image dehazing (upper) and photo retouching (lower).
        FPS is measured on 4K images using a single A100-40G.
        }
	\label{tab:quantitative}
	\resizebox{\linewidth}{!}{
        \begin{tabular}{c|ccccc|cc|cc}
            \hline
            \multirow{2}{*}{} & \multicolumn{5}{c|}{Compared Models} & \multicolumn{2}{c|}{Base Models} &  \multicolumn{2}{c}{MCT Variants} \\ \hline \hline
            ~ & MS~\cite{ren2016single} & DHN~\cite{cai2016dehazenet} & AOD~\cite{li2017aod} & GFN~\cite{ren2018gated} & MGBL~\cite{zheng2021ultra} &  GCANet &  MSBDN  & GCANet  & MSBDN \\ \hline
            \multirow{2}{*}{SOTS} & 20.31 & 21.02 & 20.27 & 23.52 & 24.50 &  25.09 &  \underline{28.56} & 25.71 & \textbf{28.70} \\
            ~ & 0.862 & 0.881 & 0.864 & 0.915 & 0.920 &  0.923 &  \textbf{0.966} & 0.927 & \underline{0.962} \\ \hline
            \multirow{2}{*}{HazeRD} & 15.35 & 15.42 & 15.44 & 14.62 & 16.06 &  16.69 &  16.23  & \textbf{17.19} & \underline{16.81} \\ 
            ~ & 0.634 & 0.622 & 0.660 & 0.580 & 0.794 &  \underline{0.825} &  0.805  & 0.810 & \textbf{0.840} \\ 
            \hline
            FPS & 13.6 & 14.8 & 41.6 & 3.0 & \underline{120.8} & 3.1 & 2.2 & \textbf{131.1} & 35.9 \\
            \hline \hline
            ~ & WB~\cite{hu2018exposure} & HDR~\cite{gharbi2017deep}   & UPE~\cite{wang2019underexposed}  & LPF~\cite{moran2020deeplpf}   & LPTN~\cite{liang2021high}   &  DPED &  DPE  & DPED  & DPE \\ \hline
            \multirow{2}{*}{Paired} & - & 23.15  & 23.24 & 24.48 & 23.86 & 24.11 & 24.14 & \underline{24.73} & \textbf{25.10} \\
            ~ & -  & 0.918 & 0.893 & 0.887 & 0.885 & 0.886 & 0.934 & \underline{0.936} & \textbf{0.941} \\ \hline
            \multirow{2}{*}{Unpaired} & 18.57  & 21.63 & 21.59 & 21.34 & 22.02 & 22.29 & 20.92  & \underline{22.81} & \textbf{23.09} \\
            ~ & 0.701  & 0.885  & 0.884 & 0.866 & 0.879 & 0.884 & 0.854  & \underline{0.902} & \textbf{0.905} \\  \hline
            FPS & 0.13 & 14.3 & 15.9 & 2.3 & 37.9 & 2.5 & 10.8 & \textbf{181.1} & \underline{162.4} \\
            \hline
        \end{tabular}
	} 
\end{table}

We quantitatively compare the performance on image dehazing and photo retouching because the images from these two tasks have corresponding ground truth to compute PSNR and SSIM.
In addition to the base models and their MCT variants, we trained some representative compared models.
Table~\ref{tab:quantitative} shows the results.
Note that HDRNet and DUPE use the open source Tensorflow~\cite{abadi2016tensorflow} implementation, while the others use PyTorch framework. 

For image dehazing, it can be seen that GCANet and MSBDN are powerful models, which are significantly better than the other compared models on the SOTS dataset.
And the MCT variants can achieve comparable or even better performance than the base models.
MSBDN performs overfitting and shows a significant performance degradation on the HazeRD dataset.
In contrast, its MCT variant has a significantly higher SSIM, which indicates that more image detail is retained.
For photo retouching, DPED and DPE are not state-of-the-art methods.
But their MCT variants outperform the base models and compared models since the curve-based methods are in line with the image retouching process.
The DPED's low SSIM is due to the small receptive field that cannot extract image contrast information effectively.
Finally, DPE performs poorly in the unpaired training setting, which may be because the CycleGAN's training strategy is not suitable for it.

In terms of runtime, FCN-based methods are significantly slower than slice-based methods, even for lightweight networks such as AODNet and LPTN. 
Since we set $H_d = W_d = 32$ for MCT-DPED, it runs faster than MCT-DPE.
Note that DUPE and HDRNet are both much slower than MCT-DPE, which is mainly due to the inefficient open-source implementations.
In our experiments, they can all reach about 180 FPS using 3D LUTs.

\begin{table}[t]
    \renewcommand\arraystretch{1.5}
    \footnotesize
    \centering
    \caption{
        Quantitative comparison (FID / user study) of the photorealistic I2I translation.
        The percentage of user study results indicates the preferred model outputs out of 95 responses.
        Lower is better for FID, and higher is better for the user study.
    }
    \label{tab:unpaired}%
    \vspace{-0.5cm}
    \begin{center}
        \resizebox{\linewidth}{!}{
            \begin{tabular}{l|c|c|c|c}
                \hline
                             & \texttt{day2dusk}                   & \texttt{dusk2day}        & \texttt{summer2autumn}               & \texttt{autumn2summer} \\
                \hline
                \hline
                CycleGAN     & \, 89.00 / 32.6\% \, & \, 94.17 / 48.4\% \, & \, \textbf{101.98} / 47.4\% \, & \, 100.34 / 40.0\% \\
                MCT-CycleGAN \, & \, \textbf{81.67} / \textbf{67.4\%} \, & \, \textbf{92.14} / \textbf{51.6\%} \, & \, 103.45 / \textbf{52.6\%} \, & \, \textbf{94.72} / \textbf{60.0\%}   \\
                \hline
                \hline
                UNIT         & \, 92.14 / 29.5\% \, & \, 96.66 / \textbf{55.8\%} \, & \, 105.15 / 42.1\%  \, & \, 95.18 / \textbf{50.5\%} \\
                MCT-UNIT     & \, \textbf{84.22} / \textbf{70.5\%} \, & \, \textbf{93.14} / 44.2\% \, & \, \textbf{103.43} / \textbf{57.9\%} \, & \, \textbf{91.35} /  49.5\%     \\
                \hline
            \end{tabular}%
        }
    \end{center}
    \vspace{-0.25cm}
\end{table}

Table~\ref{tab:unpaired} shows the quantitative comparison results of the photorealistic I2I translation.
We first compare the methods using Fréchet Inception Distance (FID)~\cite{heusel2017gans}. 
Unlike the usual experimental setup, all models translate HR images and then downsample the output images to $256 \times 256$ to compute the FID.
Although these tasks are not difficult I2I translation problems, since the input images are usually high-resolution, the base models are often not performing as well as their MCT variants.
We also released questionnaires to colleagues who were not involved in this work to conduct a small user study.
Specifically, 19 users participated in this experiment, and we provided 5 randomly selected samples for each task.
Overall, most users recognized the ability of MCT to retain high-frequency information.
In particular, for the \texttt{day2dusk} task, most users felt that the output of the MCT variants was much better than the base models.

\section{Discussion}

\noindent \textbf{Contributions.}
In this paper, we propose to modify the network's output layer for the I2I translation task.
The network is extended to predict the output pixels for the input pixel's neighboring pixels.
Since the pixels lost during downsampling are the neighboring pixels of the pixels that remain, the modified network can receive LR images to predict the mapping for process HR images.
For the adversarial training, we introduced two additional training strategies to stabilize the training.
Experimental results show that it can perform comparable or even better than the conventional models but translate 4K images in real-time on various photorealistic I2I translation tasks.

\noindent \textbf{Limitations.}
MCT is a trade-off between translation capability and speed, and it cannot be applied to the more difficult I2I translation tasks.
For tasks that the I2I translation process greatly changes the shape and texture of the objects in the image (\emph{i.e.}, high-frequency information), such as \texttt{dog2cat}, MCT is helpless.
In future research, we hope to improve its capabilities further to make it be applied to more I2I translation tasks.

\clearpage

\bibliographystyle{splncs04}
\bibliography{egbib}

\title{\large Multi-Curve Translator for High-Resolution Photorealistic Image Translation---Appendix} 
\titlerunning{Multi-Curve Translator}
\author{}
\authorrunning{Song et al.}
\institute{}

\maketitle

\renewcommand\thesection{\Alph{section}}

\section{Implementation Details}

\subsection{I2I Translation}

We use LSGAN~\cite{mao2017least} and PatchGAN~\cite{isola2017image} to train CycleGAN~\cite{zhu2017unpaired}, UNIT~\cite{liu2017unsupervised}, and their MCT variants.
The objective function of LSGAN is:
\begin{equation}
    \begin{aligned}
    \min _{D} \mathcal{L}_{LSGAN}(D) &=\frac{1}{2} \mathbb{E}_{\boldsymbol{y} \sim p(\boldsymbol{y})}\left[(D(\boldsymbol{y})-b)^{2}\right] \\
    &+\frac{1}{2} \mathbb{E}_{\boldsymbol{x} \sim p(\boldsymbol{x})}\left[(D(G(\boldsymbol{x}))-a)^{2}\right] \\
    \min _{G} \mathcal{L}_{LSGAN}(G) &=\frac{1}{2} \mathbb{E}_{\boldsymbol{x} \sim p(\boldsymbol{x})}\left[(D(G(\boldsymbol{x}))-c)^{2}\right],
    \end{aligned}
\end{equation}
where $a=c=1$ and $b=0$.
Both CycleGAN and UNIT contain two discriminators ($D_\mathcal{X}$ and $D_\mathcal{Y}$) and two generators ($G_{\mathcal{X} \rightarrow \mathcal{Y}}$ and $G_{\mathcal{Y} \rightarrow \mathcal{X}}$).
For UNIT, a generator can be further divided into an encoder $E$ and a generator $G$, then $G_{\mathcal{X} \rightarrow \mathcal{Y}} = G_\mathcal{Y}(E_\mathcal{X}(x))$ and $G_{\mathcal{Y} \rightarrow \mathcal{X}} = G_\mathcal{X}(E_\mathcal{Y}(y))$.
Note that the process of sampling latent codes is omitted here.
For simplicity, we denote the adversarial loss when training the generator $G$ as $\mathcal{L}_{gan}$.

CycleGAN and UNIT both contain the cycle-consistency loss~\cite{zhu2017unpaired,kim2017learning}, which is formulated as:
\begin{equation}
    \begin{aligned}
    \mathcal{L}_{cyc} & =\mathbb{E}_{x \sim p(x)}\left[\|G_{\mathcal{Y} \rightarrow \mathcal{X}}(G_{\mathcal{X} \rightarrow \mathcal{Y}}(x))-x\|_{1}\right] \\
    & +\mathbb{E}_{y \sim p(y)}\left[\|G_{\mathcal{X} \rightarrow \mathcal{Y}}(G_{\mathcal{Y} \rightarrow \mathcal{X}}(y))-y\|_{1}\right].
    \end{aligned}
\end{equation}

We also employ the identity loss~\cite{taigman2016unsupervised}.
For CycleGAN, it is formulated as:
\begin{equation}
    \begin{aligned}
    \mathcal{L}_{idt}&=\mathbb{E}_{y \sim p(y)}\left[\|G_{\mathcal{X} \rightarrow \mathcal{Y}}(y)-y\|_{1}\right]\\
    &+\mathbb{E}_{x \sim p(x)}\left[\|G_{\mathcal{Y} \rightarrow \mathcal{X}}(x)-x\|_{1}\right]
    \end{aligned}
\end{equation}
UNIT's identity loss is the reconstruction loss that is formulated as:
\begin{equation}
    \begin{aligned}
    \mathcal{L}_{idt}&=\mathbb{E}_{y \sim p(y)}\left[\|G_\mathcal{Y}(E_\mathcal{Y}(y))-y\|_{1}\right]\\
    &+\mathbb{E}_{x \sim p(x)}\left[\|G_\mathcal{X}(E_\mathcal{X}(x))-x\|_{1}\right]
    \end{aligned}
\end{equation}

Besides, UNIT also adds a KL divergence loss to penalize deviation of the distribution of the latent code $z_{\mathcal{X}}=E_\mathcal{X}(x)$ and $z_{\mathcal{Y}}=E_\mathcal{Y}(y)$ from the prior distribution, denoted as:
\begin{equation}
    \begin{aligned}
    \mathcal{L}_{KL} &=\operatorname{KL}\left(q_{\mathcal{X}}\left(z_{\mathcal{X}} | x\right) \| p_{\eta}(z)\right) \\
    &+ \mathrm{KL}\left(q_{\mathcal{Y}}\left(z_{\mathcal{Y}} | y\right) \| p_{\eta}(z)\right)
    \end{aligned}
\end{equation}
where the prior distribution $p_{\eta}(z)$ is a zero-mean Gaussian $p_{\eta}(z)=\mathcal{N}(z | 0, I)$.
Note the KL terms also penalize the latent codes deviating from the prior distribution in the cycle-reconstruction stream.
When training the MCT variants, we further add $\mathcal{L}_{\mathrm{reg}}$ stated in the main paper:
\begin{equation}
    \mathcal{L}_{reg} = \left\lVert G_b^{y \rightarrow x}(G_m^{x \rightarrow y}(x)) - x \right\rVert _1.
\end{equation}

The CycleGAN's full objective is:
\begin{equation}
    \mathcal{L} = \mathcal{L}_{gan} + \lambda_1 \mathcal{L}_{idt} + \lambda_2 \mathcal{L}_{cyc}.
\end{equation}
where $\lambda_1=5$ and $\lambda_2=10$.
Then the MCT-CycleGAN's full objective is:
\begin{equation}
    \mathcal{L} = \mathcal{L}_{gan} + \lambda_1 \mathcal{L}_{idt} + \lambda_2 \mathcal{L}_{cyc} + \lambda_3 \mathcal{L}_{reg}.
\end{equation}
where $\lambda_1=5$, $\lambda_2=10$, and $\lambda_3=1$.

The UNIT's full objective is:
\begin{equation}
    \mathcal{L} = \lambda_0 \mathcal{L}_{gan} + \lambda_1 \mathcal{L}_{idt} + \lambda_2 \mathcal{L}_{cyc} + \lambda_3 \mathcal{L}_{KL}.
\end{equation}
where $\lambda_0=10$, $\lambda_1=\lambda_2=100$, and $\lambda_3=0.1$.
And MCT-UNIT's full objective is:
\begin{equation}
    \mathcal{L} = \lambda_0 \mathcal{L}_{gan} + \lambda_1 \mathcal{L}_{idt} + \lambda_2 \mathcal{L}_{cyc} + \lambda_3 \mathcal{L}_{KL} + \lambda_4 \mathcal{L}_{reg}.
\end{equation}
where $\lambda_0=10$, $\lambda_1=\lambda_2=100$, $\lambda_3=0.1$ and $\lambda_4=10$.

We use the Adam optimizer~\cite{kingma2015adam} to train the models with a mini-batch size of 1. 
The base models were trained from scratch with a learning rate of $2 \times 10^{-4}$. 
We keep the same learning rate for the first 100 epochs and linearly decay the rate to zero over the next 100 epochs. 
The MCT variants load the base models' weights but also use an initial learning rate of $2 \times 10^{-4}$.
We finetune them with half of the epochs, \emph{i.e.}, keep the same learning rate for the first 50 epochs and linearly decay the rate to zero over the next 50 epochs. 

\subsection{Style Transfer}

AdaIN~\cite{huang2017arbitrary} encodes the style image and the content image using a pre-trained VGG-19~\cite{simonyan2014very} up to \texttt{relu4\_1}.
Let VGG-19 be $E$, the style image be $s$, and the content image be $c$. 
Then the transformed feature maps are:
\begin{equation}
    t=\operatorname{AdaIN}(E(c), E(s)).
\end{equation}
The goal of AdaIN is to train a decoder $G$ to reconstruct $t$ into a stylized image.
AdaIN employs content loss as:
\begin{equation}
    \mathcal{L}_{c}=\|E(G(t))-t\|_{2}
\end{equation}
Let the mean of the feature maps be $\mu$ and the variance be $\sigma$, AdaIN's style loss is:
\begin{equation}
    \begin{aligned}
    \mathcal{L}_{s}&=\sum_{i=1}^{L}\left\|\mu\left(\phi_{i}(G(t))\right)-\mu\left(\phi_{i}(s)\right)\right\|_{2} \\
    & + \sum_{i=1}^{L}\left\|\sigma\left(\phi_{i}(G(t))\right)-\sigma\left(\phi_{i}(s)\right)\right\|_{2}
    \end{aligned}
\end{equation}
where each $\phi_{i}$ denotes a layer in VGG-19 used to compute the style loss (\texttt{relu1\_1}, \texttt{relu2\_1}, \texttt{relu3\_1}, \texttt{relu4\_1}). 
Besides, we add a gradient loss $\mathcal{L}_g$ to prompt the preservation of the geometric structure as:
\begin{equation}
    \mathcal{L}_g = \left\lVert \nabla_h G(c) - \nabla_h c \right\rVert _2^2 + \left\lVert \nabla_v G(c) - \nabla_v c \right\rVert _2^2,
\end{equation}
where $\nabla_h$ ($\nabla_v$) denotes the gradient operator along the horizontal (vertical) direction.

We employs a weighted combination of the content loss $\mathcal{L}_c$, the style loss $\mathcal{L}_s$, and the gradient loss $\mathcal{L}_g$ to train AdaIN and MCT-AdaIN:
\begin{equation}
    \mathcal{L} = \mathcal{L}_c + \lambda_1 \mathcal{L}_s + \lambda_2 \mathcal{L}_g,
\end{equation}
where $\lambda_1 = 1$ and $\lambda_2 = 100$.

The key to WCT$^2$~\cite{yoo2019photorealistic} is the network architecture and the procedure of performing WCT~\cite{li2017universal}, while its training scheme is simple.
WCT$^2$ also uses the pre-trained VGG-19 as the encoder and aims to train a decoder to reconstruct the encoder's feature map into the input image.
WCT$^2$ uses the $\mathcal{L}_2$ reconstruction loss and the additional feature Gram matching loss with the encoder to train the decoder.
For simplicity, we only keep the reconstruction loss because the Gram matching loss is not necessary.
The $\mathcal{L}_2$ reconstruction loss is formulated as:
\begin{equation}
    \mathcal{L}=\|G(E(c))-c\|_{2}.
\end{equation}
Then its MCT variant's loss function is:
\begin{equation}
    \mathcal{L} = \|G_m(E(c))-c\|_{2} + \|G_b(E(c))-c\|_{2}.
\end{equation}

The challenge of extending WCT$^2$ to its MCT variant is that it is extremely easy to fall into a collapse solution, even if we constrain the base output.
Specifically, the curves represented by the parameter maps obtained by summing the two outputs of the MCT variant are always identity mapping, even if we perform WCT on the intermediate feature maps.
For this reason, we need to further prevent the slicing operation from leaking low-frequency information from the input to the output.
First, we perform the grayscale operation on the HR image.
Then the HR image contains only one channel, so the corresponding number of curves is changed from 9 to 3 ($C=3M$).
When performing stylization, we further match the HR image's brightness with the style image's brightness to prevent the brightness of the HR image from being retained in the output image, which can be expressed as $\tilde{c} = c - \mu(c) + \mu(s)$.
Note that we do not grayscale and brightness-align the LR images.

We use the Adam optimizer to train the models with a mini-batch size of 8. 
The base models were trained from scratch with a learning rate of $1 \times 10^{-4}$ for $1 \times 10^{5}$ iterations.
Their MCT variants load the base models' weights, and are trained with a learning rate of $1 \times 10^{-4}$ for $5 \times 10^{4}$ iterations.

\subsection{Image Dehazing}

We use only the $\mathcal{L}_1$ loss function to train GCANet~\cite{chen2019gated}, MSBDN~\cite{dong2020multi}, and their MCT variants:
\begin{equation}
    \mathcal{L}=\|G(x)-y\|_{1}.
\end{equation}

All models are trained from scratch using the Adam optimizer with the cosine annealing strategy~\cite{loshchilov2016sgdr} but not warm restarts.
For all models, we set the initial learning rate to $1 \times 10^{-4}$.
For GCANet and MCT-GCANet, we set mini-batch size to 56 and train them for $500$ epochs.
For MSBDN and MCT-MSBDN, we set mini-batch size to 28 and train them for $300$ epochs.

\subsection{Photo Retouching}

Since color mapping and receptive field are critical for photo retouching, we set $H_d = W_d = 32$, $M=32$ for MCT-DPED and set $M = 16$ and $p=8$ for MCT-DPE.

For paired training, we also use the $\mathcal{L}_1$ loss function to train DPED~\cite{ignatov2017dslr} and DPE~\cite{chen2018deep}.
All models are trained from scratch using the Adam optimizer with the cosine annealing strategy but not warm restarts.
And they are trained from scratch with a learning rate of $1 \times 10^{-4}$ for $100$ epochs.
Since the memory consumptions of these models are different, we train them using different mini-batch sizes.
For DPED, we set the mini-batch size to 16.
For DPED-MCT and DPE, we set the mini-batch size to 64.
For DPE-MCT, we set the mini-batch size to 32.

For unpaired training, we employ the earlier described CycleGAN's training scheme.
For DPED, all training hyperparameters are not modified.
For DPE, we removed the identity loss because we found that it would cause the DPE to fall into a poor solution.

\section{Limitations \& Future Works}

MCT is a flexible framework that minimizes the effort of modifying the base model into an MCT variant.
For a new I2I translation task, we only need to modify the output layer of the existing model to extend it to its MCT variant.
Besides, we can not only load the pre-training weights of the base model but also use the training strategy of the base model directly without modifying any hyperparameters.
Unfortunately, there are limitations to the tasks for which MCT is applicable.

The first is the tasks to which MCT can be applied.
We assume that if the I2I translation process only slightly changes the shape and texture of the objects in the image (\emph{i.e.}, high-frequency information), then a mapping responsible for local regions in the image domain can be learned to approximate this translation process.
For tasks that do not satisfy our assumptions, such as \texttt{dog2cat}, MCT is helpless.
The reason is that MCT is realized by interpolation to apply the LR parameter maps to transform HR images, so the transformation of high-frequency information is spatially smooth.
However, for some I2I translation tasks with large shape changes, the transformation of high-frequency information is highly unsmoothed spatially.
We have tried to introduce spatial support similar to KPN~\cite{mildenhall2018burst}, but no visible improvement has been achieved.
It may mean that improving the translator's capability to translate high-frequency information without significantly increasing the number of parameters and the computational cost is a non-trivial task.
And it is the focus of our future research.

In addition to this, MCT is not always plug-and-play.
For CycleGAN and UNIT, we need to constrain the base output to prevent the MCT variants from falling into collapse solutions.
For WCT$^2$, we need to grayscale and brightness-align the HR images to reduce further the low-frequency information flowing from the input to the output through the slicing operation.
It is necessary to develop more generalized training strategies to reduce further the difficulty of extending the base model to their MCT variants.

Finally, although the MCT variants are significantly faster compared to their base models, they may still be computationally heavy for edge devices due to the high computational cost of some base models (\emph{e.g.} CycleGAN).
For this, we may need to introduce the model compression approach or design lightweight network architectures to lower the computational cost of the backbone network.
In addition, the runtime of the lookup table should not be ignored since the memory bandwidth of the edge device can limit the speed of slicing operations.
We plan to find a better dimensionality reduction operation than image downsampling and revise the slicing operation to reduce computational and memory access costs.

\section{More Experimental Results}

We included the table of FID in the main paper, and we expanded it to FID / KID $\times$ 100 in Table~\ref{tab:fid}.
This experiment provides a rough indication of each method's translation capability for HR images, and the results are for reference only since the FID and KID are not suitable for evaluating HR images.

We expand the user study to style transfer in Table~\ref{tab:user}.
Although the results of AdaIN preserve almost no high-frequency information, most users still feel that the quality of MCT-AdaIN is inferior.
AdaIN is an artistic style transfer method, and its fixed VGG encoder without skip connection severely loses details.
This property conflicts with the property of MCT to retain high-frequency information, making MCT-AdaIN's results unattractive.

We also explore not reintroducing high-frequency information, but using super-resolution to predict high-frequency information.
We conducted experiments on image retouching (see Table~\ref{tab:tua}).
This method does not perform well, especially in terms of SSIM, and we found that little lost  high-frequency information can be restored in the output.

We then quantitatively compare the translation capabilities of the state-of-the-art lightweight I2I translation network and the MCT variants using supervised learning-based tasks.
Table~\ref{tab:lptn_compare} shows the quantitative comparison between LPTN~\cite{liang2021high} and MCT-DPE on the photo retouching.
It can be seen that LPTN is significantly inferior to MCT-DPE in both speed and performance.
It is worth mentioning that LPTN only achieves 23.09 dB on the SOTS dataset, 2.62 dB lower than MCT-GCANet, and 5.61 dB lower than MCT-MSBDN for image dehazing.

Considering that MCT is a curve-based method, we further compare MCT with other curve-based methods.
Previous curve-based methods were employed only on image retouching.
Since GleNet~\cite{kim2020global} was tested on downsampled images and provided an empty repository, we re-implemented GleNet's GEN (LEN is not real-time).
We train CURL~\cite{moran2021curl} in the unpaired setting because no previous works reported the result.
We did not train StarEnhancer~\cite{song2021starenhancer} in the unpaired setting because it is a multi-style method that is non-trivial to extend.
Table~\ref{tab:quantitative2} shows the comparison.
The global transformation introduces inductive biases practical for image retouching, making the previous curve-based methods prevent overfitting and run fast.
In contrast, DPE as a base model does not effectively aggregate global information.

\begin{figure*}[htp]
    \vspace{2mm}
    \centering
    \includegraphics[width=1.0\textwidth]{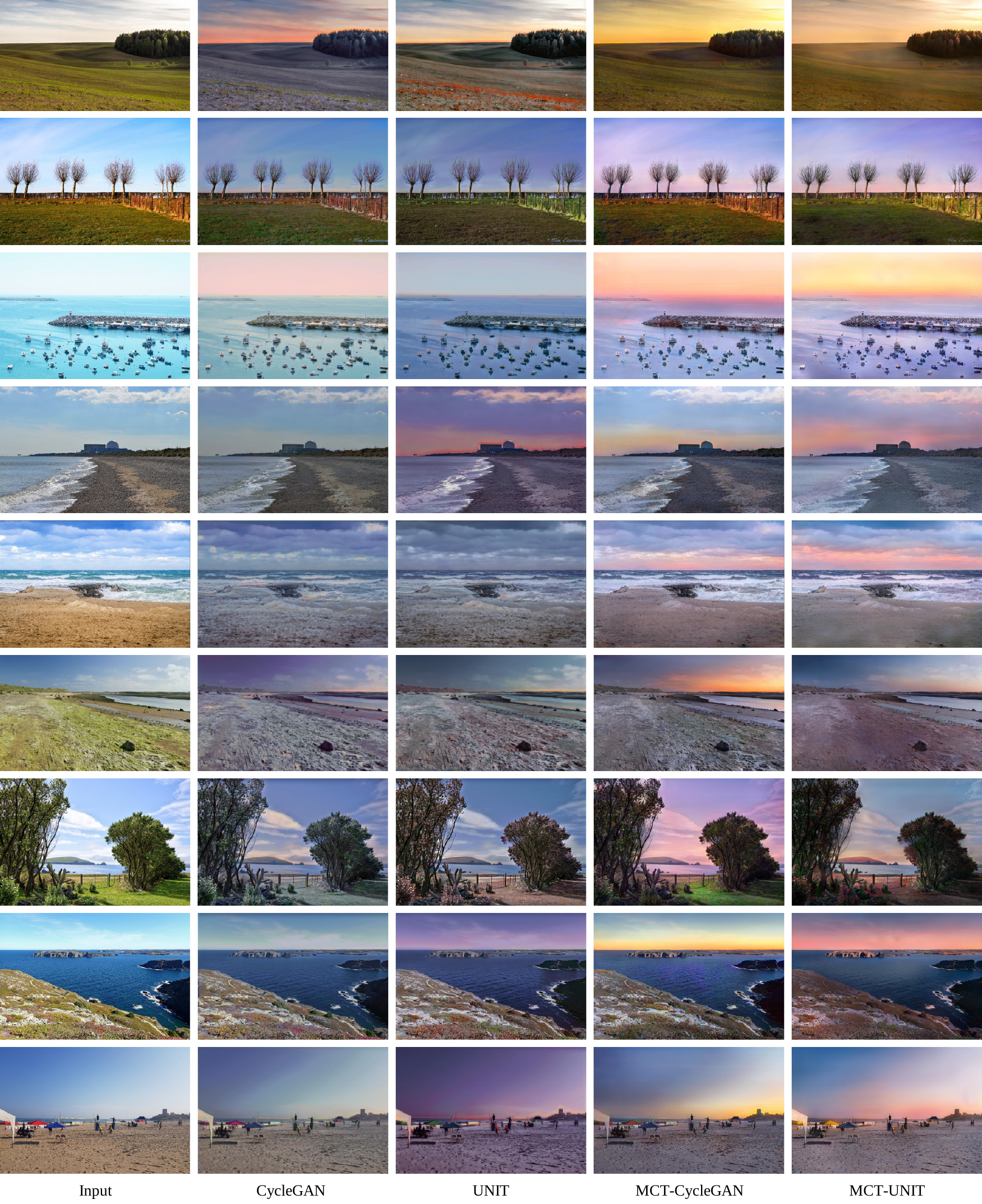}
    \caption{
        Qualitative comparison of \texttt{day2dusk}.
    }
    \label{fig:exp1}
\end{figure*}

\begin{figure*}[htp]
    \vspace{2mm}
    \centering
    \includegraphics[width=1.0\textwidth]{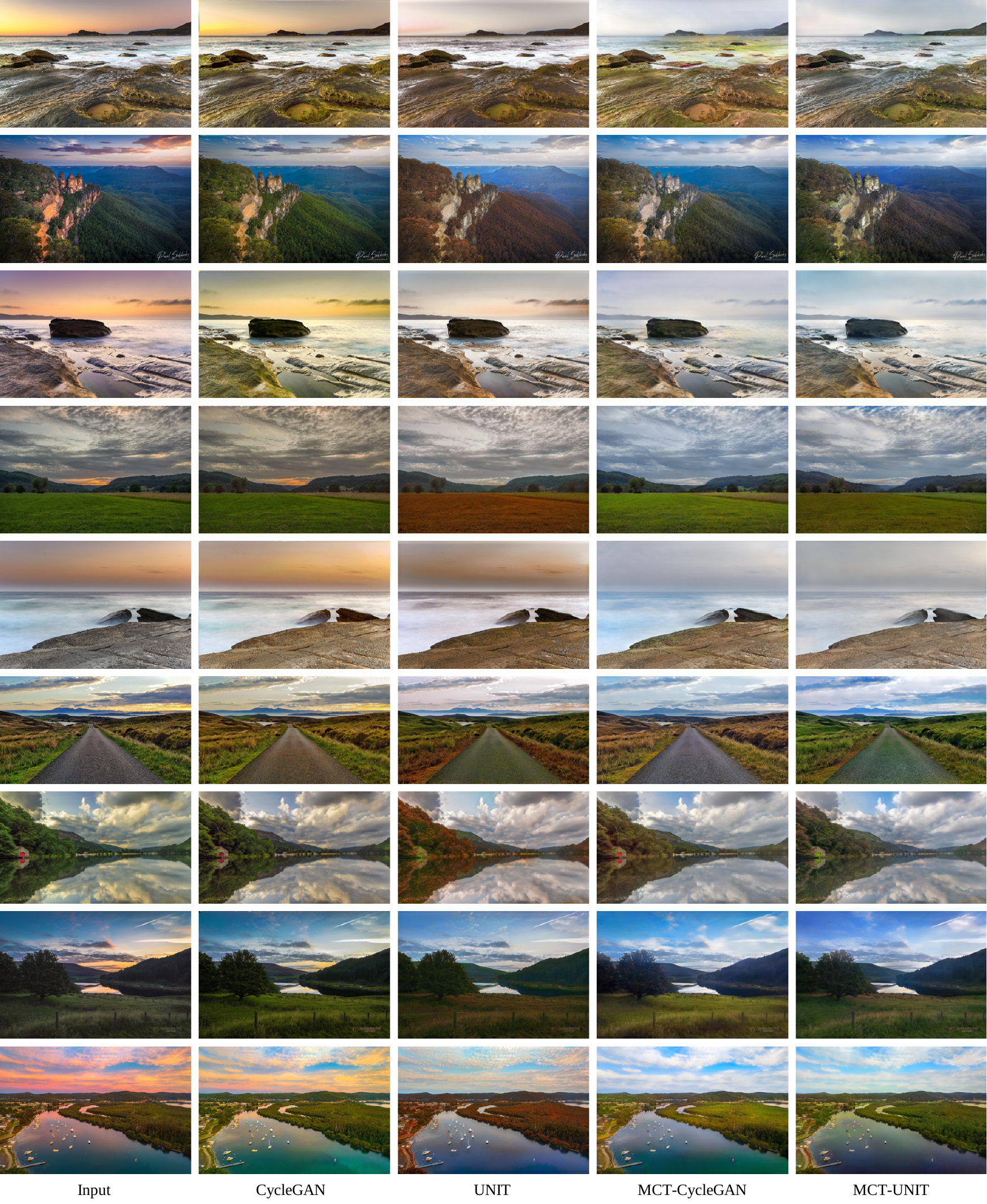}
    \caption{
        Qualitative comparison of \texttt{dusk2day}.
    }
    \label{fig:exp2}
\end{figure*}

\begin{figure*}[htp]
    \vspace{2mm}
    \centering
    \includegraphics[width=1.0\textwidth]{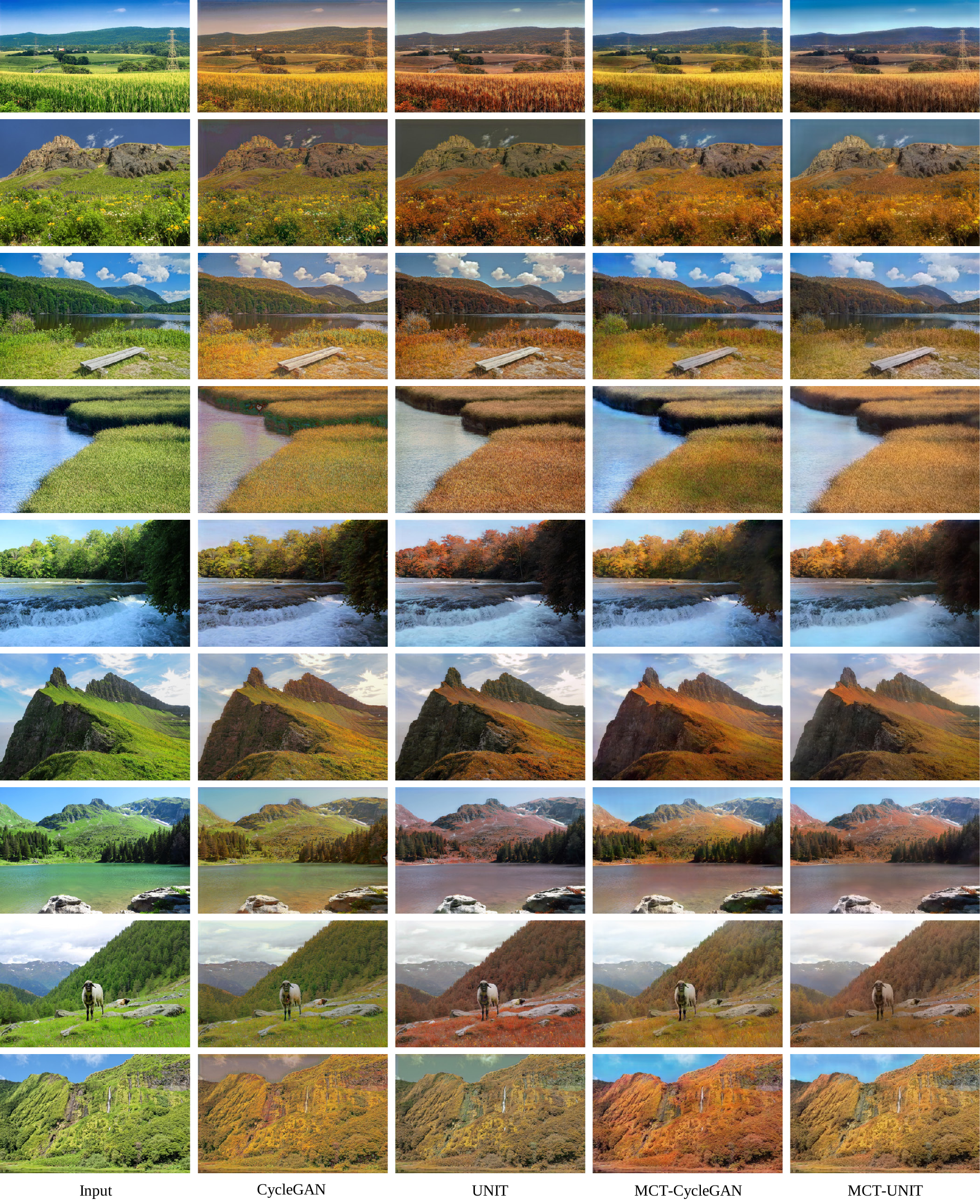}
    \caption{
        Qualitative comparison of \texttt{summer2autumn}.
    }
    \label{fig:exp3}
\end{figure*}

\begin{figure*}[htp]
    \vspace{2mm}
    \centering
    \includegraphics[width=1.0\textwidth]{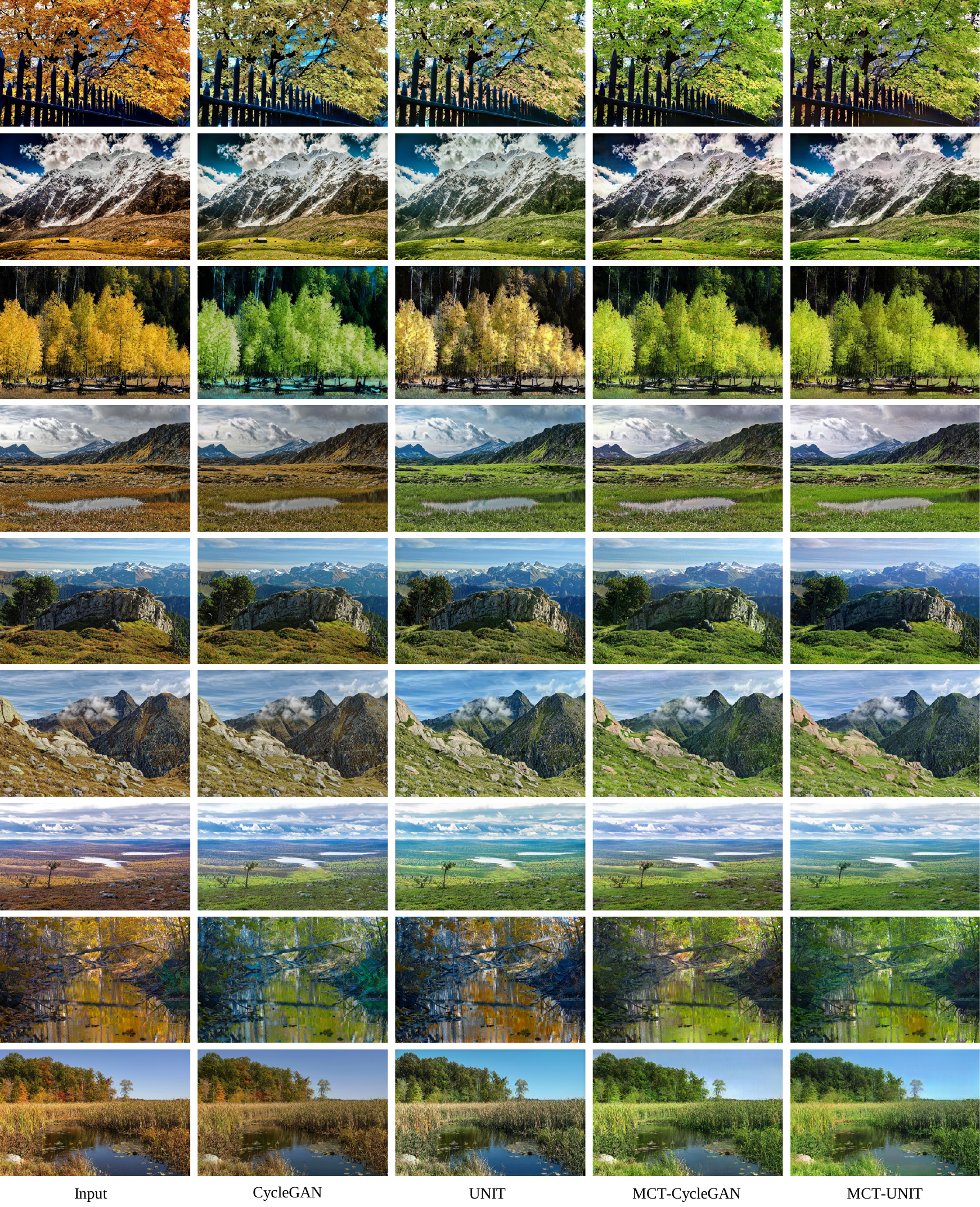}
    \caption{
        Qualitative comparison of \texttt{autumn2summer}.
    }
    \label{fig:exp4}
\end{figure*}

\begin{figure*}[htp]
    \vspace{5mm}
    \centering
    \includegraphics[width=1.0\textwidth]{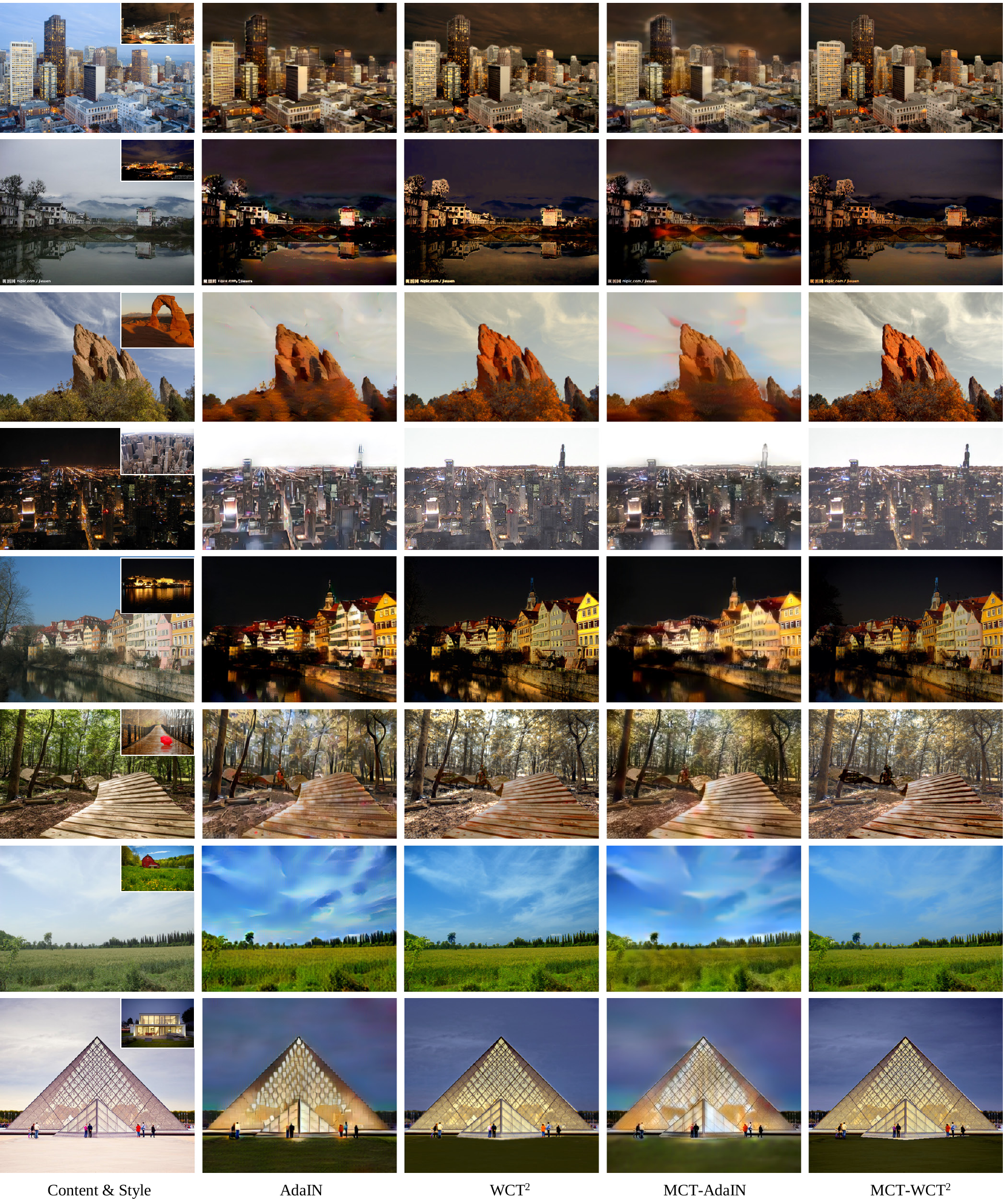}
    \caption{
        Qualitative comparison of style transfer with paired segmentation label maps.
    }
    \label{fig:exp5}
\end{figure*}

\begin{figure*}[htp]
    \vspace{3mm}
    \centering
    \includegraphics[width=1.0\textwidth]{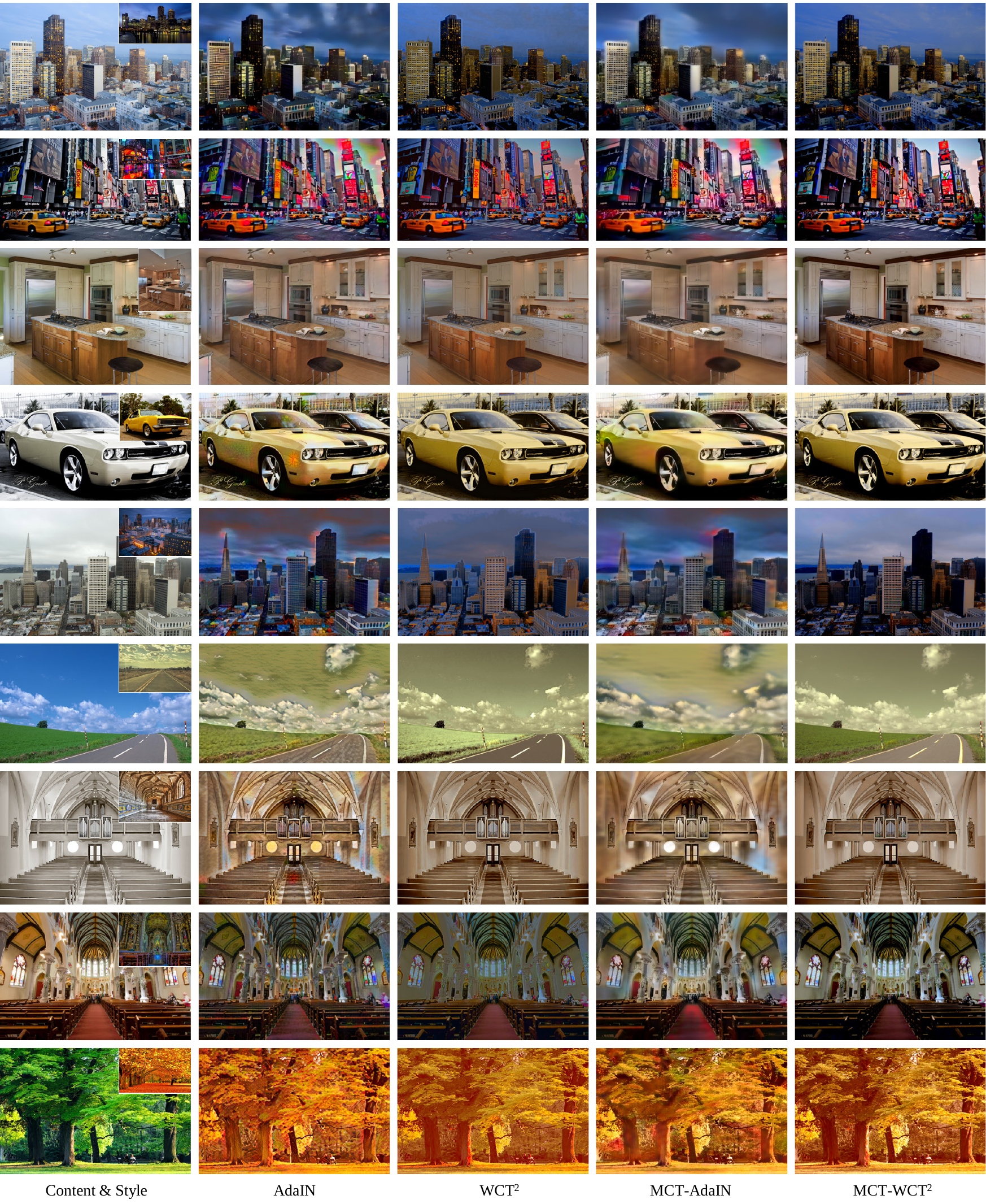}
    \caption{
        Qualitative comparison of style transfer without segmentation label maps.
    }
    \label{fig:exp6}
\end{figure*}

\begin{figure*}[htp]
    \vspace{5mm}
    \centering
    \includegraphics[width=1.0\textwidth]{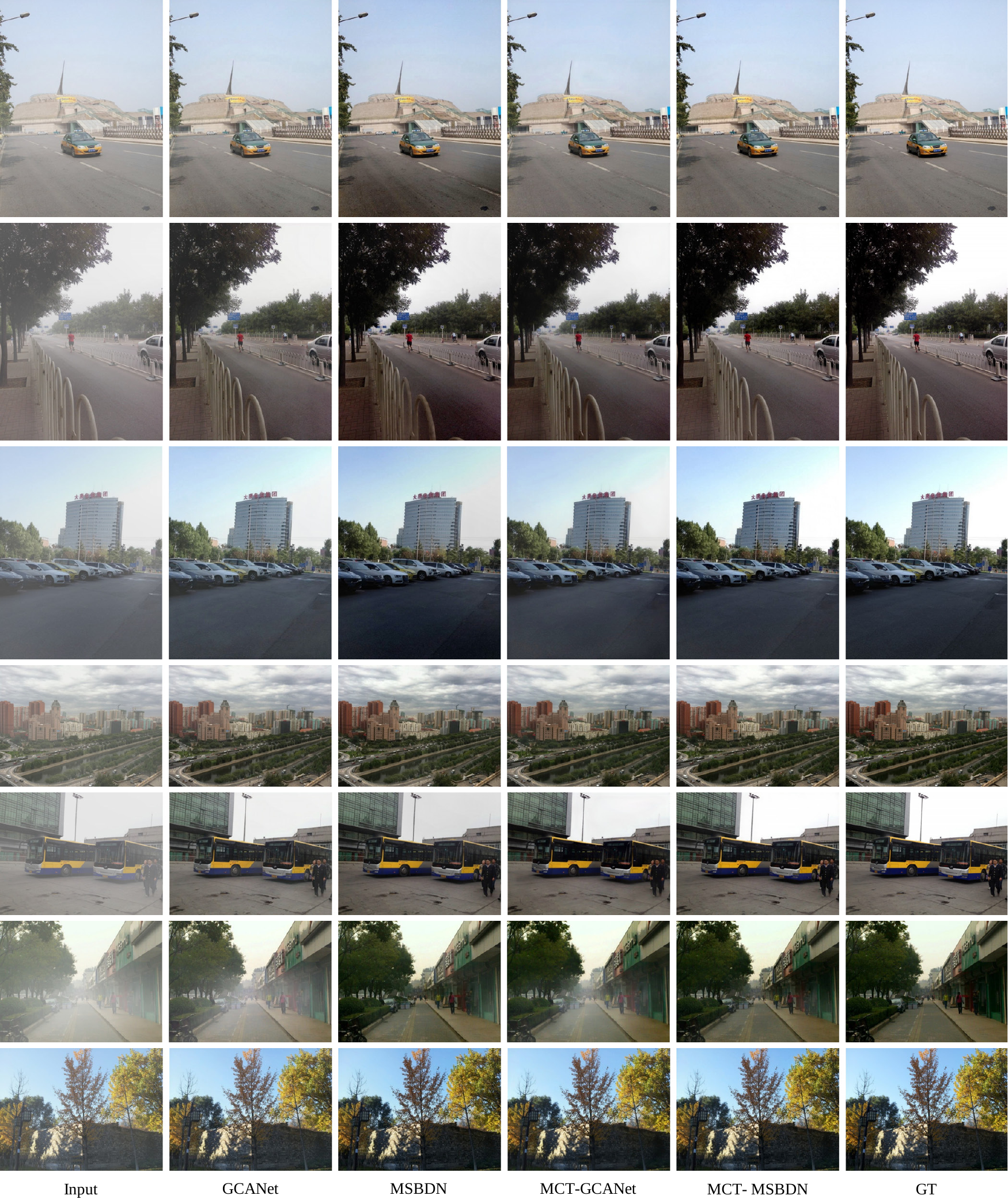}
    \caption{
        Qualitative comparison of image dehazing on SOTS dataset~\cite{li2018benchmarking}.
    }
    \label{fig:exp7}
\end{figure*}

\begin{figure*}[htp]
    \vspace{5mm}
    \centering
    \includegraphics[width=1.0\textwidth]{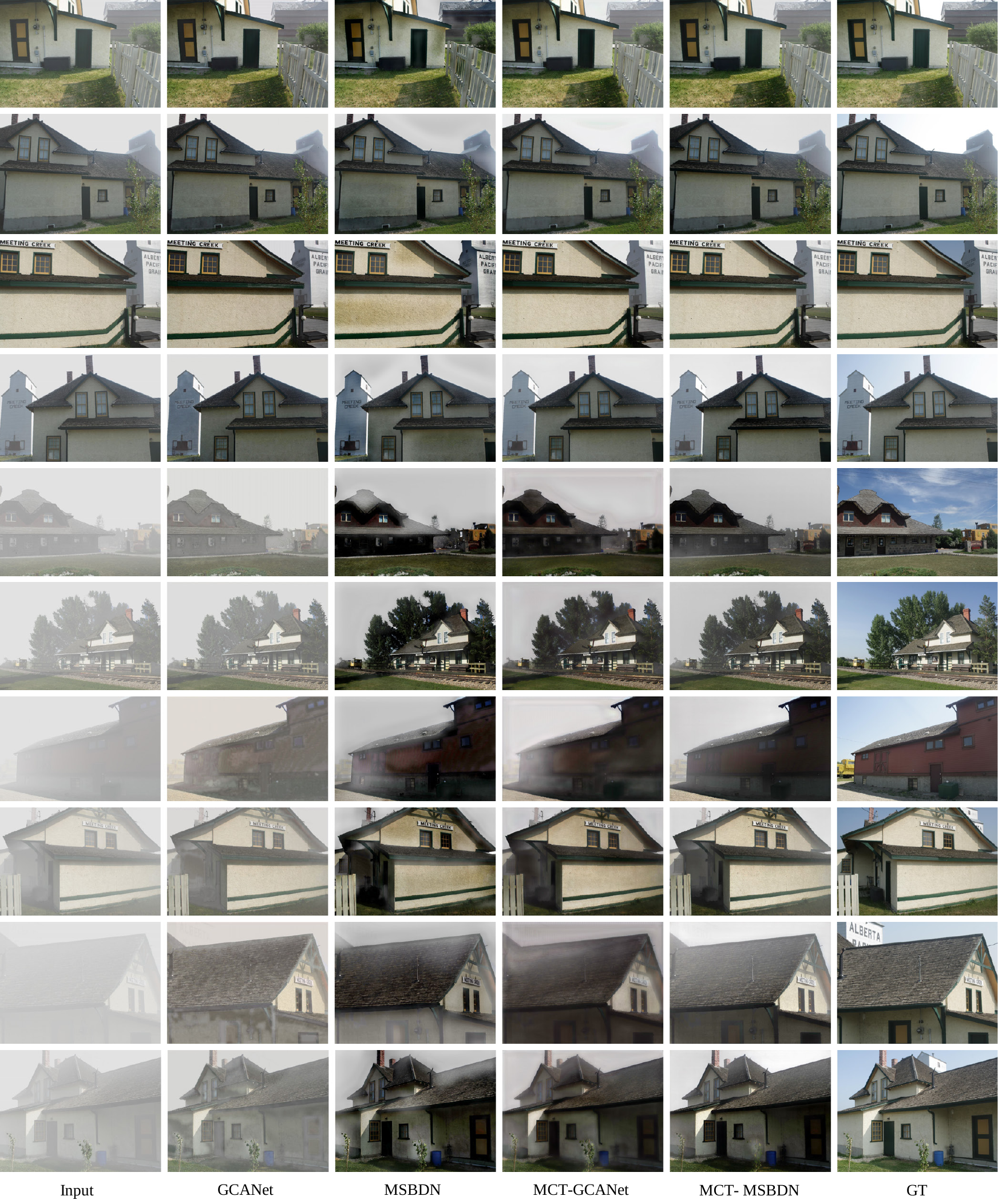}
    \caption{
        Qualitative comparison of image dehazing on HazeRD dataset~\cite{zhang2017hazerd}.
    }
    \label{fig:exp8}
\end{figure*}

\begin{figure*}[htp]
    \vspace{1cm}
    \centering
    \includegraphics[width=1.0\textwidth]{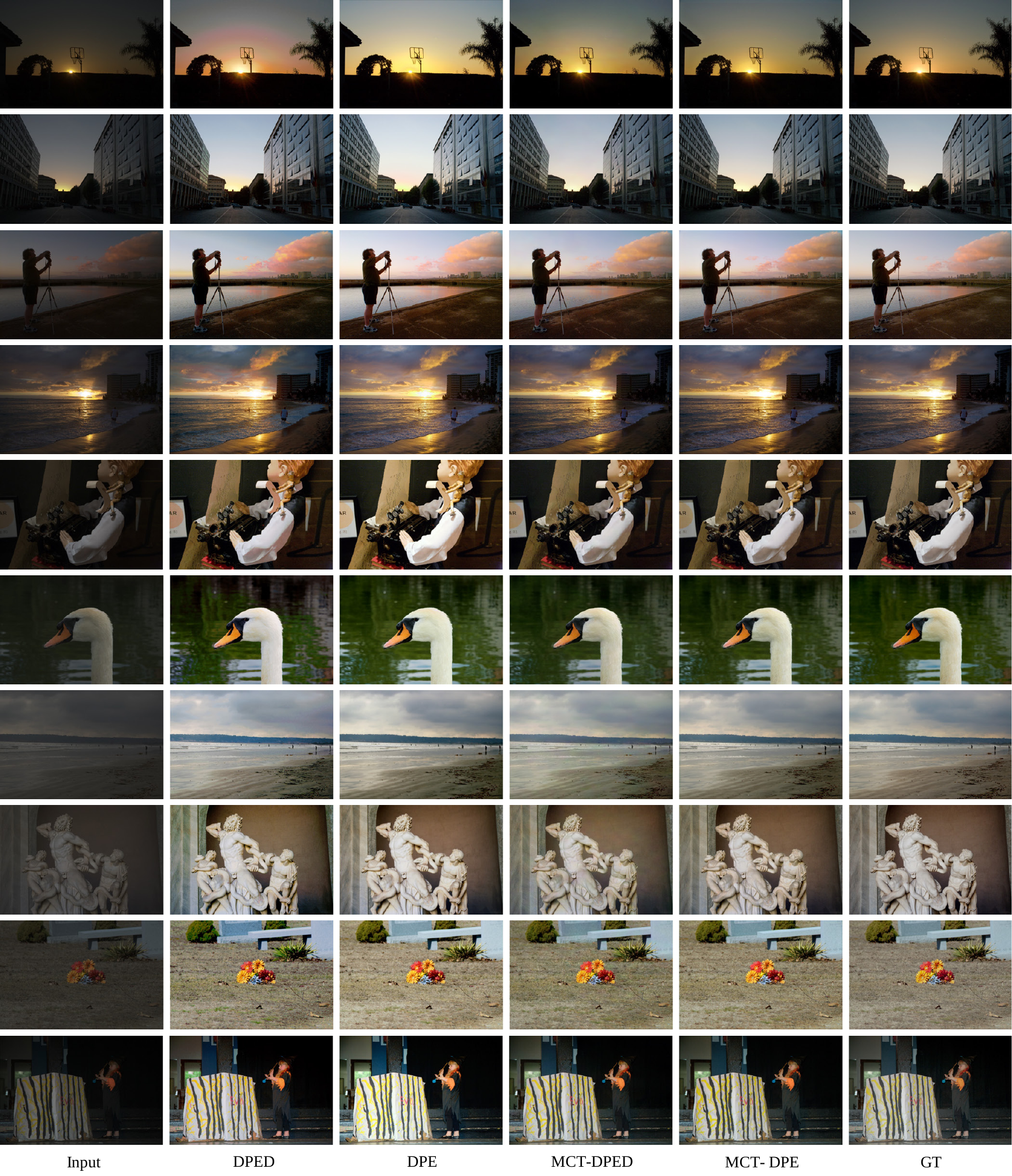}
    \caption{
        Qualitative comparison of photo retouching with paired training.
    }
    \label{fig:exp9}
\end{figure*}

\begin{figure*}[htp]
    \vspace{1cm}
    \centering
    \includegraphics[width=1.0\textwidth]{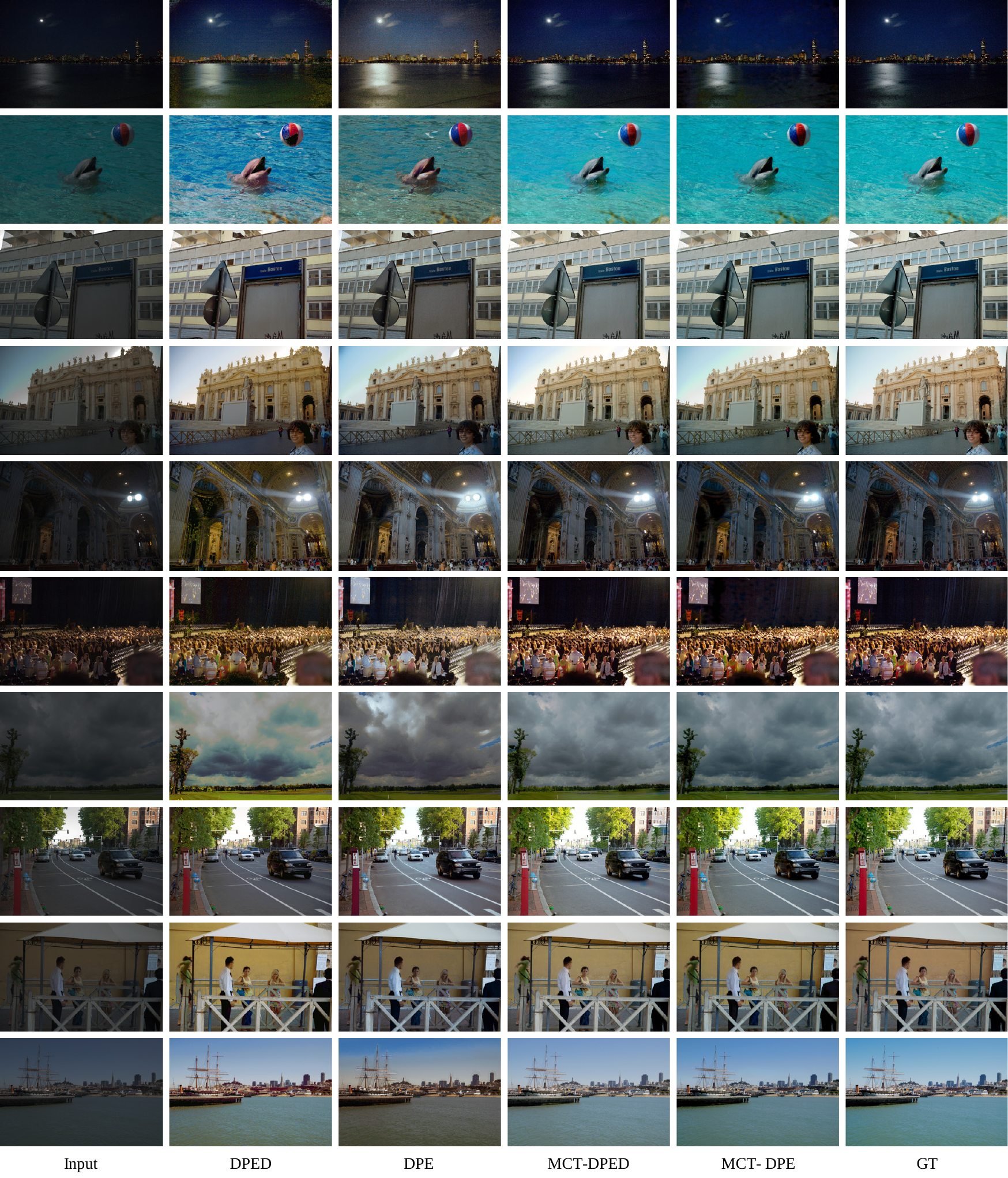}
    \caption{
        Qualitative comparison of photo retouching with unpaired training.
    }
    \label{fig:exp10}
\end{figure*}

Table~\ref{tab:speed} illustates more runtime comparison results. 
Figures~\ref{fig:exp1}-\ref{fig:exp10} show more qualitative comparison results.
Readers can generate more test results using the provided code and pre-trained models.

\begin{figure}[htp]
    \centering
    \includegraphics[width=1.0\textwidth]{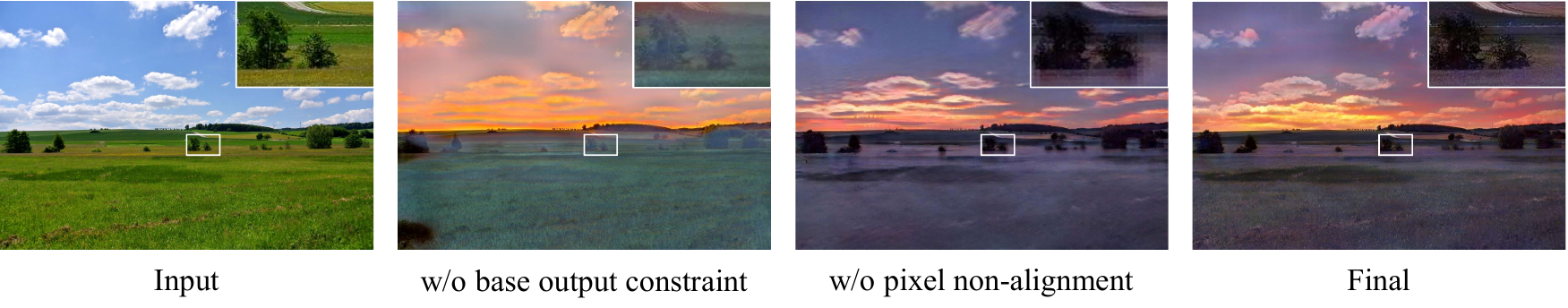}
    \caption{
        Ablation study on \texttt{day2dusk}.
        The second image is the result without constraining the base output during training.
        The third image is the result without the pixel unalignment training strategy.
        The last image is the result using our proposed full training scheme.
    }
    \label{fig:ablation}
    \vspace{-0.5cm}
\end{figure}

We finally visualize the effectiveness of the two training strategies.
Fig.~\ref{fig:ablation} shows a special case when training MCT-CycleGAN on \texttt{day2dusk}.
If we train MCT-CycleGAN without constraining the base output, it may fall into a poor solution.
In contrast, imposing constraints on the base output makes the backbone network responsible for low-frequency information and medium-frequency information, leaving only the high-frequency information lost during downsampling to be taken care of by the slicing operation.
When we do not use the pixel unaligned training strategy, the output image of MCT may lose high-frequency information.
Unlike increasing the weight of cycle-consistency loss, a pixel unaligned training strategy causes the slicing operation to focus more on high-frequency information.
Note that although the output of MCT is blurred at this point, it still contains more high-frequency information than the upsampled base output due to the curve slicing operation.

\begin{table}[htp]
    \centering
    \caption{
       Quantitative comparison (FID / KID $\times$ 100) of the photorealistic I2I translation.
       Lower is better.
    }
    \label{tab:fid}%
    \vspace{-0.5cm}
    \begin{center}
       \resizebox{0.85\linewidth}{!}{
          \begin{tabular}{l|c|c|c|c}
             \hline
                          & \texttt{day2dusk}                     & \texttt{dusk2day}         & \texttt{summer2autumn}                   & \texttt{autumn2summer}               \\
             \hline
             CycleGAN     & 89.00 / 1.14                         & 94.17 / \underline{1.69} & \textbf{101.98} / \underline{1.45} & 100.34 / 1.65                  \\
             UNIT         & 92.14 / 1.38                         & 96.66 / \textbf{1.58}    & 105.15 / 1.54                      & 95.18 / \underline{1.50}       \\
             \hline
             MCT-CycleGAN & \textbf{81.67} / \textbf{0.67}       & \textbf{92.14} / 1.75    & 103.45 / 1.56                      & \underline{94.72} / 1.55       \\
             MCT-UNIT     & \underline{84.22} / \underline{1.01} & \underline{93.14} / 1.72 & \underline{103.43} / \textbf{1.44} & \textbf{91.35} / \textbf{1.48} \\
             \hline
          \end{tabular}%
       }
    \end{center}
    \vspace{-0.5cm}
 \end{table}%

  \begin{table}[htp]
     \footnotesize
     \centering
     \caption{
        User study results.
        The percentage indicates the preferred model outputs out of 95 responses.
        Note that \texttt{d2d} means \texttt{day2dusk}, \texttt{s2a} means \texttt{summer2autumn}, and \texttt{M} means Mask.
     }
     \label{tab:user}
     \vspace{-0.5cm}
     \begin{center}
        \resizebox{0.78\linewidth}{!}{
           \begin{tabular}{c|cc|cc|cc|cc}
              \hline
              Methods      & \multicolumn{2}{c|}{CycleGAN}     & \multicolumn{2}{c|}{UNIT}         & \multicolumn{2}{c|}{AdaIN}        & \multicolumn{2}{c}{WCT$^2$}                                                                                                                                                 \\
              \hline
              Task         & \texttt{d2d}                      & \texttt{s2a}                      & \texttt{d2d}                      & \texttt{s2a}                      & w/ \texttt{M}                     & w/o \texttt{M}                    & w/ \texttt{M}                     & w/o \texttt{M}           \\
              \hline
              \hline
              Base         &  32.6\%           &  47.4\%           &  29.5\%           &  42.1\%           &  \textbf{84.2\%}  &  \textbf{89.5\%}  &  \textbf{54.7}\%  &  42.1\%          \\
              \textbf{MCT} &  \textbf{67.4\%}  &  \textbf{52.6\%}  &  \textbf{70.5\%}  &  \textbf{57.9\%}  &  15.8\%           &  10.5\%           &  45.3\%           &  \textbf{57.9\%} \\
              \hline
           \end{tabular}
        }
     \end{center}
     \vspace{-0.5cm}
  \end{table}

  \begin{table}[htp]
     \footnotesize
     \centering
     \caption{
        Quantitative comparison of photo retouching in the unpaired setting.
        The FPS is tested using a single A100.
        TU means Translation-Upsampling (use EDSR-LIIF~\cite{chen2021learning} for scale-arbitrary).
     }
     \label{tab:tua}
     \vspace{-0.5cm}
     \begin{center}
        \resizebox{0.9\linewidth}{!}{
           \begin{tabular}{c|c|c|c}
              \hline
              \multirow{2}*{Methods} & 480p                                    & 1080p                                            & original                                         \\
              \cline{2-4}
                                     & PSNR / SSIM / FPS                       & PSNR / SSIM / FPS                                & PSNR / SSIM / FPS                                \\
              \hline
              \hline
              DPE                    & 21.07 / 0.861 / \textbf{284.7}          & 21.02 / 0.859 / 43.7                             & 20.92 / 0.854 / 10.8                             \\
              TU-DPE                 & 19.47 / 0.730 / 13.7                    & 18.83 / 0.673 / 2.2                              & 18.53 / 0.654 / 0.6                              \\
              \textbf{MCT-DPE}       & \textbf{23.40} / \textbf{0.903} / 271.6 & \textbf{23.31} / \textbf{0.903} / \textbf{269.9} & \textbf{23.09} / \textbf{0.905} / \textbf{153.8} \\
              \hline
           \end{tabular}
        }
     \end{center}
     \vspace{-0.5cm}
  \end{table}

 \begin{table}[htp]
    \renewcommand\arraystretch{1.5}
    \footnotesize
    \centering
    \caption{Quantitative comparison on the FiveK dataset in the unpaired setting.
    The FPS is tested on A100 with batch size $=1$.}
    \label{tab:lptn_compare}
    \vspace{-0.5cm}
    \begin{center}
       \resizebox{0.8\linewidth}{!}{
       \begin{tabular}{c|c|c|c|c|c|c|c|c|c}
          \hline
          \multirow{2}*{Methods}&\multicolumn{3}{c|}{480p}&\multicolumn{3}{c|}{1080p}&\multicolumn{3}{c}{original}\\
          \cline{2-10}
          & PSNR & SSIM & FPS & PSNR & SSIM & FPS & PSNR & SSIM & FPS\\
          \hline
          \hline
          LPTN ($ L=3 $) & 22.12 & 0.878 & 248.9  & 22.09 & 0.883 & 188.4 & 22.02 & 0.879 & 37.9 \\
          \textbf{MCT-DPE} & \textbf{23.40} & \textbf{0.903} & \textbf{271.6}  & \textbf{23.31} & \textbf{0.903} & \textbf{269.9} & \textbf{23.09} & \textbf{0.905} & \textbf{153.8} \\
          \hline
       \end{tabular}
       }
    \end{center}
    \vspace{-0.5cm}
 \end{table}	

\begin{table}[htp]
   \footnotesize
   \centering
   \caption{
      Quantitative comparison of photo retouching.
      FPS is measured on 4K images using a single A100.
      Note that some results are replicated from~\cite{li2020flexible,song2021starenhancer}.
   }
   \label{tab:quantitative2}
   \begin{center}
    \vspace{-0.5cm}
    \resizebox{0.58\linewidth}{!}{
         \begin{tabular}{c|cc|cc|c}
            \hline
            \multirow{2}*{Methods}                   & \multicolumn{2}{c|}{Paired} & \multicolumn{2}{c|}{Unpaired} & \multirow{2}*{FPS}                                         \\
            \cline{2-5}
                                                     & PSNR                        & SSIM                          & PSNR               & SSIM              &                   \\
            \hline
            \hline
            FlexiCurve~\cite{li2020flexible}         & 23.97                  & 0.910                    & 22.12         & 0.860       & 83.3         \\
            CURL~\cite{moran2021curl}                & 24.20                  & 0.880                    & 21.62              & 0.873             & 3.4               \\
            GEN~\cite{kim2020global}                 & 24.91                       & 0.937                         & \underline{22.73}  & \underline{0.902} & \textbf{364.3}    \\
            StarEnhancer~\cite{song2021starenhancer} & \textbf{25.29}         & \textbf{0.943}           & -                  & -                 & \underline{242.1} \\
            \hline
            \textbf{MCT-DPE}                         & \underline{25.10}           & \underline{0.941}             & \textbf{23.09}     & \textbf{0.905}    & 153.8             \\
            \hline
         \end{tabular}
      }
   \end{center}
   \vspace{-0.5cm}
\end{table}

\newpage

\begin{table*}[htp]
    \centering
    \caption{Runtime comparison of the base models and their MCT variants.}
    \resizebox{\linewidth}{!}{
        \begin{tabular}{|c|c|ccccccccc|}
            \hline
            \multirow{2}{*}{Method} & \multicolumn{1}{c|}{\multirow{2}{*}{Hardware}} & \multicolumn{9}{c|}{Resolution} \\
        \cline{3-11}          & \multicolumn{1}{c|}{} & \multicolumn{1}{c|}{256×256} & \multicolumn{1}{c|}{360×360} & \multicolumn{1}{c|}{512×512} & \multicolumn{1}{c|}{1280×720} & \multicolumn{1}{c|}{1920×1080} & \multicolumn{1}{c|}{2560×1440} & \multicolumn{1}{c|}{3840×2160} & \multicolumn{1}{c|}{6000×4000} & 7680×4320 \\
            \hline
            \hline
            \multirow{8}{*}{CycleGAN} & A100-40G & \cellcolor[rgb]{ .776,  .937,  .808}\textcolor[rgb]{ 0,  .38,  0}{238.4 } & \cellcolor[rgb]{ .776,  .937,  .808}\textcolor[rgb]{ 0,  .38,  0}{146.1 } & \cellcolor[rgb]{ .776,  .937,  .808}\textcolor[rgb]{ 0,  .38,  0}{80.7 } & 23.8  & 10.8  & 6.1   & 2.7   & \cellcolor[rgb]{ 1,  .78,  .808}\textcolor[rgb]{ .612,  0,  .024}{OOM} & \cellcolor[rgb]{ 1,  .78,  .808}\textcolor[rgb]{ .612,  0,  .024}{OOM} \\
                  & RTX 3090 & \cellcolor[rgb]{ .776,  .937,  .808}\textcolor[rgb]{ 0,  .38,  0}{138.6 } & \cellcolor[rgb]{ .776,  .937,  .808}\textcolor[rgb]{ 0,  .38,  0}{72.7 } & \cellcolor[rgb]{ .776,  .937,  .808}\textcolor[rgb]{ 0,  .38,  0}{37.1 } & 11.7  & 5.2   & 2.9   & 1.3   & \cellcolor[rgb]{ 1,  .78,  .808}\textcolor[rgb]{ .612,  0,  .024}{OOM} & \cellcolor[rgb]{ 1,  .78,  .808}\textcolor[rgb]{ .612,  0,  .024}{OOM} \\
                  & RTX 3080 & \cellcolor[rgb]{ .776,  .937,  .808}\textcolor[rgb]{ 0,  .38,  0}{130.7 } & \cellcolor[rgb]{ .776,  .937,  .808}\textcolor[rgb]{ 0,  .38,  0}{65.6 } & \cellcolor[rgb]{ .776,  .937,  .808}\textcolor[rgb]{ 0,  .38,  0}{32.8 } & 7.1   & 4.3   & 2.4   & 1.1   & \cellcolor[rgb]{ 1,  .78,  .808}\textcolor[rgb]{ .612,  0,  .024}{OOM} & \cellcolor[rgb]{ 1,  .78,  .808}\textcolor[rgb]{ .612,  0,  .024}{OOM} \\
                  & RTX 3070 & \cellcolor[rgb]{ .776,  .937,  .808}\textcolor[rgb]{ 0,  .38,  0}{77.3 } & \cellcolor[rgb]{ .776,  .937,  .808}\textcolor[rgb]{ 0,  .38,  0}{41.2 } & 21.4  & 6.0   & 2.9   & 1.7   & \cellcolor[rgb]{ 1,  .78,  .808}\textcolor[rgb]{ .612,  0,  .024}{OOM} & \cellcolor[rgb]{ 1,  .78,  .808}\textcolor[rgb]{ .612,  0,  .024}{OOM} & \cellcolor[rgb]{ 1,  .78,  .808}\textcolor[rgb]{ .612,  0,  .024}{OOM} \\
                  & RTX 3060 & \cellcolor[rgb]{ .776,  .937,  .808}\textcolor[rgb]{ 0,  .38,  0}{52.6 } & 29.4  & 15.0  & 4.4   & 2.0   & 1.1   & 0.5   & \cellcolor[rgb]{ 1,  .78,  .808}\textcolor[rgb]{ .612,  0,  .024}{OOM} & \cellcolor[rgb]{ 1,  .78,  .808}\textcolor[rgb]{ .612,  0,  .024}{OOM} \\
                  & RTX 2080Ti & \cellcolor[rgb]{ .776,  .937,  .808}\textcolor[rgb]{ 0,  .38,  0}{111.4 } & \cellcolor[rgb]{ .776,  .937,  .808}\textcolor[rgb]{ 0,  .38,  0}{52.1 } & 27.6  & 7.4   & 3.3   & 1.9   & 0.9   & \cellcolor[rgb]{ 1,  .78,  .808}\textcolor[rgb]{ .612,  0,  .024}{OOM} & \cellcolor[rgb]{ 1,  .78,  .808}\textcolor[rgb]{ .612,  0,  .024}{OOM} \\
                  & GTX 1080Ti & \cellcolor[rgb]{ .776,  .937,  .808}\textcolor[rgb]{ 0,  .38,  0}{58.1 } & 29.6  & 15.7  & 4.1   & 1.9   & 1.0   & 0.5   & \cellcolor[rgb]{ 1,  .78,  .808}\textcolor[rgb]{ .612,  0,  .024}{OOM} & \cellcolor[rgb]{ 1,  .78,  .808}\textcolor[rgb]{ .612,  0,  .024}{OOM} \\
                  & GTX 1070Ti & 17.5  & 16.7  & 8.4   & 2.3   & 1.0   & 0.6   & \cellcolor[rgb]{ 1,  .78,  .808}\textcolor[rgb]{ .612,  0,  .024}{OOM} & \cellcolor[rgb]{ 1,  .78,  .808}\textcolor[rgb]{ .612,  0,  .024}{OOM} & \cellcolor[rgb]{ 1,  .78,  .808}\textcolor[rgb]{ .612,  0,  .024}{OOM} \\
            \hline
            \multirow{8}{*}{WCT$^2$} & A100-40G & 14.2  & 13.3  & 11.1  & 4.7   & 2.1   & 1.2   & \cellcolor[rgb]{ 1,  .78,  .808}\textcolor[rgb]{ .612,  0,  .024}{OOM} & \cellcolor[rgb]{ 1,  .78,  .808}\textcolor[rgb]{ .612,  0,  .024}{OOM} & \cellcolor[rgb]{ 1,  .78,  .808}\textcolor[rgb]{ .612,  0,  .024}{OOM} \\
                  & RTX 3090 & 13.7  & 13.1  & 8.7   & 3.4   & 1.7   & 1.0   & \cellcolor[rgb]{ 1,  .78,  .808}\textcolor[rgb]{ .612,  0,  .024}{OOM} & \cellcolor[rgb]{ 1,  .78,  .808}\textcolor[rgb]{ .612,  0,  .024}{OOM} & \cellcolor[rgb]{ 1,  .78,  .808}\textcolor[rgb]{ .612,  0,  .024}{OOM} \\
                  & RTX 3080 & 12.9  & 10.2  & 7.2   & 2.9   & \cellcolor[rgb]{ 1,  .78,  .808}\textcolor[rgb]{ .612,  0,  .024}{OOM} & \cellcolor[rgb]{ 1,  .78,  .808}\textcolor[rgb]{ .612,  0,  .024}{OOM} & \cellcolor[rgb]{ 1,  .78,  .808}\textcolor[rgb]{ .612,  0,  .024}{OOM} & \cellcolor[rgb]{ 1,  .78,  .808}\textcolor[rgb]{ .612,  0,  .024}{OOM} & \cellcolor[rgb]{ 1,  .78,  .808}\textcolor[rgb]{ .612,  0,  .024}{OOM} \\
                  & RTX 3070 & 11.3  & 9.2   & 5.9   & \cellcolor[rgb]{ 1,  .78,  .808}\textcolor[rgb]{ .612,  0,  .024}{OOM} & \cellcolor[rgb]{ 1,  .78,  .808}\textcolor[rgb]{ .612,  0,  .024}{OOM} & \cellcolor[rgb]{ 1,  .78,  .808}\textcolor[rgb]{ .612,  0,  .024}{OOM} & \cellcolor[rgb]{ 1,  .78,  .808}\textcolor[rgb]{ .612,  0,  .024}{OOM} & \cellcolor[rgb]{ 1,  .78,  .808}\textcolor[rgb]{ .612,  0,  .024}{OOM} & \cellcolor[rgb]{ 1,  .78,  .808}\textcolor[rgb]{ .612,  0,  .024}{OOM} \\
                  & RTX 3060 & 10.6  & 7.2   & 4.4   & 1.4   & \cellcolor[rgb]{ 1,  .78,  .808}\textcolor[rgb]{ .612,  0,  .024}{OOM} & \cellcolor[rgb]{ 1,  .78,  .808}\textcolor[rgb]{ .612,  0,  .024}{OOM} & \cellcolor[rgb]{ 1,  .78,  .808}\textcolor[rgb]{ .612,  0,  .024}{OOM} & \cellcolor[rgb]{ 1,  .78,  .808}\textcolor[rgb]{ .612,  0,  .024}{OOM} & \cellcolor[rgb]{ 1,  .78,  .808}\textcolor[rgb]{ .612,  0,  .024}{OOM} \\
                  & RTX 2080Ti & 12.4  & 10.7  & 7.0   & 2.5   & \cellcolor[rgb]{ 1,  .78,  .808}\textcolor[rgb]{ .612,  0,  .024}{OOM} & \cellcolor[rgb]{ 1,  .78,  .808}\textcolor[rgb]{ .612,  0,  .024}{OOM} & \cellcolor[rgb]{ 1,  .78,  .808}\textcolor[rgb]{ .612,  0,  .024}{OOM} & \cellcolor[rgb]{ 1,  .78,  .808}\textcolor[rgb]{ .612,  0,  .024}{OOM} & \cellcolor[rgb]{ 1,  .78,  .808}\textcolor[rgb]{ .612,  0,  .024}{OOM} \\
                  & GTX 1080Ti & 8.9   & 6.8   & 4.3   & 1.2   & \cellcolor[rgb]{ 1,  .78,  .808}\textcolor[rgb]{ .612,  0,  .024}{OOM} & \cellcolor[rgb]{ 1,  .78,  .808}\textcolor[rgb]{ .612,  0,  .024}{OOM} & \cellcolor[rgb]{ 1,  .78,  .808}\textcolor[rgb]{ .612,  0,  .024}{OOM} & \cellcolor[rgb]{ 1,  .78,  .808}\textcolor[rgb]{ .612,  0,  .024}{OOM} & \cellcolor[rgb]{ 1,  .78,  .808}\textcolor[rgb]{ .612,  0,  .024}{OOM} \\
                  & GTX 1070Ti & 7.6   & 4.7   & 2.8   & 0.7   & \cellcolor[rgb]{ 1,  .78,  .808}\textcolor[rgb]{ .612,  0,  .024}{OOM} & \cellcolor[rgb]{ 1,  .78,  .808}\textcolor[rgb]{ .612,  0,  .024}{OOM} & \cellcolor[rgb]{ 1,  .78,  .808}\textcolor[rgb]{ .612,  0,  .024}{OOM} & \cellcolor[rgb]{ 1,  .78,  .808}\textcolor[rgb]{ .612,  0,  .024}{OOM} & \cellcolor[rgb]{ 1,  .78,  .808}\textcolor[rgb]{ .612,  0,  .024}{OOM} \\
            \hline
            \multirow{8}{*}{GCANet} & A100-40G & \cellcolor[rgb]{ .776,  .937,  .808}\textcolor[rgb]{ 0,  .38,  0}{235.1 } & \cellcolor[rgb]{ .776,  .937,  .808}\textcolor[rgb]{ 0,  .38,  0}{199.0 } & \cellcolor[rgb]{ .776,  .937,  .808}\textcolor[rgb]{ 0,  .38,  0}{95.6 } & 28.2  & 12.2  & 6.9   & 3.1   & 1.0   & \cellcolor[rgb]{ 1,  .78,  .808}\textcolor[rgb]{ .612,  0,  .024}{OOM} \\
                  & RTX 3090 & \cellcolor[rgb]{ .776,  .937,  .808}\textcolor[rgb]{ 0,  .38,  0}{229.4 } & \cellcolor[rgb]{ .776,  .937,  .808}\textcolor[rgb]{ 0,  .38,  0}{138.7 } & \cellcolor[rgb]{ .776,  .937,  .808}\textcolor[rgb]{ 0,  .38,  0}{75.7 } & 21.9  & 9.8   & 5.5   & 2.4   & \cellcolor[rgb]{ 1,  .78,  .808}\textcolor[rgb]{ .612,  0,  .024}{OOM} & \cellcolor[rgb]{ 1,  .78,  .808}\textcolor[rgb]{ .612,  0,  .024}{OOM} \\
                  & RTX 3080 & \cellcolor[rgb]{ .776,  .937,  .808}\textcolor[rgb]{ 0,  .38,  0}{227.2 } & \cellcolor[rgb]{ .776,  .937,  .808}\textcolor[rgb]{ 0,  .38,  0}{121.4 } & \cellcolor[rgb]{ .776,  .937,  .808}\textcolor[rgb]{ 0,  .38,  0}{64.7 } & 18.5  & 8.3   & 4.6   & \cellcolor[rgb]{ 1,  .78,  .808}\textcolor[rgb]{ .612,  0,  .024}{OOM} & \cellcolor[rgb]{ 1,  .78,  .808}\textcolor[rgb]{ .612,  0,  .024}{OOM} & \cellcolor[rgb]{ 1,  .78,  .808}\textcolor[rgb]{ .612,  0,  .024}{OOM} \\
                  & RTX 3070 & \cellcolor[rgb]{ .776,  .937,  .808}\textcolor[rgb]{ 0,  .38,  0}{168.6 } & \cellcolor[rgb]{ .776,  .937,  .808}\textcolor[rgb]{ 0,  .38,  0}{81.1 } & \cellcolor[rgb]{ .776,  .937,  .808}\textcolor[rgb]{ 0,  .38,  0}{43.3 } & 12.2  & 5.5   & 3.1   & \cellcolor[rgb]{ 1,  .78,  .808}\textcolor[rgb]{ .612,  0,  .024}{OOM} & \cellcolor[rgb]{ 1,  .78,  .808}\textcolor[rgb]{ .612,  0,  .024}{OOM} & \cellcolor[rgb]{ 1,  .78,  .808}\textcolor[rgb]{ .612,  0,  .024}{OOM} \\
                  & RTX 3060 & \cellcolor[rgb]{ .776,  .937,  .808}\textcolor[rgb]{ 0,  .38,  0}{117.4 } & \cellcolor[rgb]{ .776,  .937,  .808}\textcolor[rgb]{ 0,  .38,  0}{58.2 } & \cellcolor[rgb]{ .776,  .937,  .808}\textcolor[rgb]{ 0,  .38,  0}{30.8 } & 8.7   & 3.9   & 2.2   & \cellcolor[rgb]{ 1,  .78,  .808}\textcolor[rgb]{ .612,  0,  .024}{OOM} & \cellcolor[rgb]{ 1,  .78,  .808}\textcolor[rgb]{ .612,  0,  .024}{OOM} & \cellcolor[rgb]{ 1,  .78,  .808}\textcolor[rgb]{ .612,  0,  .024}{OOM} \\
                  & RTX 2080Ti & \cellcolor[rgb]{ .776,  .937,  .808}\textcolor[rgb]{ 0,  .38,  0}{205.6 } & \cellcolor[rgb]{ .776,  .937,  .808}\textcolor[rgb]{ 0,  .38,  0}{78.5 } & \cellcolor[rgb]{ .776,  .937,  .808}\textcolor[rgb]{ 0,  .38,  0}{52.0 } & 14.0  & 6.1   & 3.5   & \cellcolor[rgb]{ 1,  .78,  .808}\textcolor[rgb]{ .612,  0,  .024}{OOM} & \cellcolor[rgb]{ 1,  .78,  .808}\textcolor[rgb]{ .612,  0,  .024}{OOM} & \cellcolor[rgb]{ 1,  .78,  .808}\textcolor[rgb]{ .612,  0,  .024}{OOM} \\
                  & GTX 1080Ti & \cellcolor[rgb]{ .776,  .937,  .808}\textcolor[rgb]{ 0,  .38,  0}{90.5 } & \cellcolor[rgb]{ .776,  .937,  .808}\textcolor[rgb]{ 0,  .38,  0}{41.3 } & 20.5  & 5.3   & 2.3   & 1.3   & \cellcolor[rgb]{ 1,  .78,  .808}\textcolor[rgb]{ .612,  0,  .024}{OOM} & \cellcolor[rgb]{ 1,  .78,  .808}\textcolor[rgb]{ .612,  0,  .024}{OOM} & \cellcolor[rgb]{ 1,  .78,  .808}\textcolor[rgb]{ .612,  0,  .024}{OOM} \\
                  & GTX 1070Ti & \cellcolor[rgb]{ .776,  .937,  .808}\textcolor[rgb]{ 0,  .38,  0}{60.6 } & 28.6  & 14.2  & 3.8   & 1.7   & 0.9   & \cellcolor[rgb]{ 1,  .78,  .808}\textcolor[rgb]{ .612,  0,  .024}{OOM} & \cellcolor[rgb]{ 1,  .78,  .808}\textcolor[rgb]{ .612,  0,  .024}{OOM} & \cellcolor[rgb]{ 1,  .78,  .808}\textcolor[rgb]{ .612,  0,  .024}{OOM} \\
            \hline
            \multirow{8}{*}{DPE} & A100-40G & \cellcolor[rgb]{ .776,  .937,  .808}\textcolor[rgb]{ 0,  .38,  0}{410.6 } & \cellcolor[rgb]{ .776,  .937,  .808}\textcolor[rgb]{ 0,  .38,  0}{403.5 } & \cellcolor[rgb]{ .776,  .937,  .808}\textcolor[rgb]{ 0,  .38,  0}{317.9 } & \cellcolor[rgb]{ .776,  .937,  .808}\textcolor[rgb]{ 0,  .38,  0}{94.8 } & \cellcolor[rgb]{ .776,  .937,  .808}\textcolor[rgb]{ 0,  .38,  0}{43.7 } & 24.4  & 10.8  & 3.6   & 2.6  \\
                  & RTX 3090 & \cellcolor[rgb]{ .776,  .937,  .808}\textcolor[rgb]{ 0,  .38,  0}{400.3 } & \cellcolor[rgb]{ .776,  .937,  .808}\textcolor[rgb]{ 0,  .38,  0}{390.6 } & \cellcolor[rgb]{ .776,  .937,  .808}\textcolor[rgb]{ 0,  .38,  0}{279.1 } & \cellcolor[rgb]{ .776,  .937,  .808}\textcolor[rgb]{ 0,  .38,  0}{84.2 } & \cellcolor[rgb]{ .776,  .937,  .808}\textcolor[rgb]{ 0,  .38,  0}{39.1 } & 21.9  & 9.9   & \cellcolor[rgb]{ 1,  .78,  .808}\textcolor[rgb]{ .612,  0,  .024}{OOM} & \cellcolor[rgb]{ 1,  .78,  .808}\textcolor[rgb]{ .612,  0,  .024}{OOM} \\
                  & RTX 3080 & \cellcolor[rgb]{ .776,  .937,  .808}\textcolor[rgb]{ 0,  .38,  0}{396.4 } & \cellcolor[rgb]{ .776,  .937,  .808}\textcolor[rgb]{ 0,  .38,  0}{357.3 } & \cellcolor[rgb]{ .776,  .937,  .808}\textcolor[rgb]{ 0,  .38,  0}{252.2 } & \cellcolor[rgb]{ .776,  .937,  .808}\textcolor[rgb]{ 0,  .38,  0}{74.5 } & \cellcolor[rgb]{ .776,  .937,  .808}\textcolor[rgb]{ 0,  .38,  0}{34.5 } & 19.1  & \cellcolor[rgb]{ 1,  .78,  .808}\textcolor[rgb]{ .612,  0,  .024}{OOM} & \cellcolor[rgb]{ 1,  .78,  .808}\textcolor[rgb]{ .612,  0,  .024}{OOM} & \cellcolor[rgb]{ 1,  .78,  .808}\textcolor[rgb]{ .612,  0,  .024}{OOM} \\
                  & RTX 3070 & \cellcolor[rgb]{ .776,  .937,  .808}\textcolor[rgb]{ 0,  .38,  0}{382.0 } & \cellcolor[rgb]{ .776,  .937,  .808}\textcolor[rgb]{ 0,  .38,  0}{315.4 } & \cellcolor[rgb]{ .776,  .937,  .808}\textcolor[rgb]{ 0,  .38,  0}{175.8 } & \cellcolor[rgb]{ .776,  .937,  .808}\textcolor[rgb]{ 0,  .38,  0}{51.2 } & 23.5  & 13.1  & \cellcolor[rgb]{ 1,  .78,  .808}\textcolor[rgb]{ .612,  0,  .024}{OOM} & \cellcolor[rgb]{ 1,  .78,  .808}\textcolor[rgb]{ .612,  0,  .024}{OOM} & \cellcolor[rgb]{ 1,  .78,  .808}\textcolor[rgb]{ .612,  0,  .024}{OOM} \\
                  & RTX 3060 & \cellcolor[rgb]{ .776,  .937,  .808}\textcolor[rgb]{ 0,  .38,  0}{372.1 } & \cellcolor[rgb]{ .776,  .937,  .808}\textcolor[rgb]{ 0,  .38,  0}{231.3 } & \cellcolor[rgb]{ .776,  .937,  .808}\textcolor[rgb]{ 0,  .38,  0}{127.6 } & \cellcolor[rgb]{ .776,  .937,  .808}\textcolor[rgb]{ 0,  .38,  0}{37.2 } & 17.0  & 9.5   & 4.2   & \cellcolor[rgb]{ 1,  .78,  .808}\textcolor[rgb]{ .612,  0,  .024}{OOM} & \cellcolor[rgb]{ 1,  .78,  .808}\textcolor[rgb]{ .612,  0,  .024}{OOM} \\
                  & RTX 2080Ti & \cellcolor[rgb]{ .776,  .937,  .808}\textcolor[rgb]{ 0,  .38,  0}{389.3 } & \cellcolor[rgb]{ .776,  .937,  .808}\textcolor[rgb]{ 0,  .38,  0}{345.7 } & \cellcolor[rgb]{ .776,  .937,  .808}\textcolor[rgb]{ 0,  .38,  0}{195.8 } & \cellcolor[rgb]{ .776,  .937,  .808}\textcolor[rgb]{ 0,  .38,  0}{60.0 } & 26.9  & 15.3  & 6.7   & \cellcolor[rgb]{ 1,  .78,  .808}\textcolor[rgb]{ .612,  0,  .024}{OOM} & \cellcolor[rgb]{ 1,  .78,  .808}\textcolor[rgb]{ .612,  0,  .024}{OOM} \\
                  & GTX 1080Ti & \cellcolor[rgb]{ .776,  .937,  .808}\textcolor[rgb]{ 0,  .38,  0}{378.8 } & \cellcolor[rgb]{ .776,  .937,  .808}\textcolor[rgb]{ 0,  .38,  0}{238.8 } & \cellcolor[rgb]{ .776,  .937,  .808}\textcolor[rgb]{ 0,  .38,  0}{137.5 } & \cellcolor[rgb]{ .776,  .937,  .808}\textcolor[rgb]{ 0,  .38,  0}{37.5 } & 16.7  & 9.3   & 4.1   & \cellcolor[rgb]{ 1,  .78,  .808}\textcolor[rgb]{ .612,  0,  .024}{OOM} & \cellcolor[rgb]{ 1,  .78,  .808}\textcolor[rgb]{ .612,  0,  .024}{OOM} \\
                  & GTX 1070Ti & \cellcolor[rgb]{ .776,  .937,  .808}\textcolor[rgb]{ 0,  .38,  0}{273.3 } & \cellcolor[rgb]{ .776,  .937,  .808}\textcolor[rgb]{ 0,  .38,  0}{153.7 } & \cellcolor[rgb]{ .776,  .937,  .808}\textcolor[rgb]{ 0,  .38,  0}{84.2 } & 22.2  & 10.0  & 5.5   & \cellcolor[rgb]{ 1,  .78,  .808}\textcolor[rgb]{ .612,  0,  .024}{OOM} & \cellcolor[rgb]{ 1,  .78,  .808}\textcolor[rgb]{ .612,  0,  .024}{OOM} & \cellcolor[rgb]{ 1,  .78,  .808}\textcolor[rgb]{ .612,  0,  .024}{OOM} \\
            \hline
            \hline
            \multirow{8}{*}{MCT-CycleGAN} & A100-40G & \cellcolor[rgb]{ .776,  .937,  .808}\textcolor[rgb]{ 0,  .38,  0}{173.5 } & \cellcolor[rgb]{ .776,  .937,  .808}\textcolor[rgb]{ 0,  .38,  0}{172.3 } & \cellcolor[rgb]{ .776,  .937,  .808}\textcolor[rgb]{ 0,  .38,  0}{171.5 } & \cellcolor[rgb]{ .776,  .937,  .808}\textcolor[rgb]{ 0,  .38,  0}{169.6 } & \cellcolor[rgb]{ .776,  .937,  .808}\textcolor[rgb]{ 0,  .38,  0}{168.9 } & \cellcolor[rgb]{ .776,  .937,  .808}\textcolor[rgb]{ 0,  .38,  0}{153.1 } & \cellcolor[rgb]{ .776,  .937,  .808}\textcolor[rgb]{ 0,  .38,  0}{116.0 } & \cellcolor[rgb]{ .776,  .937,  .808}\textcolor[rgb]{ 0,  .38,  0}{64.2 } & \cellcolor[rgb]{ .776,  .937,  .808}\textcolor[rgb]{ 0,  .38,  0}{51.1 } \\
                  & RTX 3090 & \cellcolor[rgb]{ .776,  .937,  .808}\textcolor[rgb]{ 0,  .38,  0}{116.1 } & \cellcolor[rgb]{ .776,  .937,  .808}\textcolor[rgb]{ 0,  .38,  0}{115.9 } & \cellcolor[rgb]{ .776,  .937,  .808}\textcolor[rgb]{ 0,  .38,  0}{114.6 } & \cellcolor[rgb]{ .776,  .937,  .808}\textcolor[rgb]{ 0,  .38,  0}{111.1 } & \cellcolor[rgb]{ .776,  .937,  .808}\textcolor[rgb]{ 0,  .38,  0}{103.5 } & \cellcolor[rgb]{ .776,  .937,  .808}\textcolor[rgb]{ 0,  .38,  0}{95.1 } & \cellcolor[rgb]{ .776,  .937,  .808}\textcolor[rgb]{ 0,  .38,  0}{77.7 } & \cellcolor[rgb]{ .776,  .937,  .808}\textcolor[rgb]{ 0,  .38,  0}{47.6 } & \cellcolor[rgb]{ .776,  .937,  .808}\textcolor[rgb]{ 0,  .38,  0}{39.3 } \\
                  & RTX 3080 & \cellcolor[rgb]{ .776,  .937,  .808}\textcolor[rgb]{ 0,  .38,  0}{108.9 } & \cellcolor[rgb]{ .776,  .937,  .808}\textcolor[rgb]{ 0,  .38,  0}{108.1 } & \cellcolor[rgb]{ .776,  .937,  .808}\textcolor[rgb]{ 0,  .38,  0}{106.4 } & \cellcolor[rgb]{ .776,  .937,  .808}\textcolor[rgb]{ 0,  .38,  0}{100.7 } & \cellcolor[rgb]{ .776,  .937,  .808}\textcolor[rgb]{ 0,  .38,  0}{93.1 } & \cellcolor[rgb]{ .776,  .937,  .808}\textcolor[rgb]{ 0,  .38,  0}{84.7 } & \cellcolor[rgb]{ .776,  .937,  .808}\textcolor[rgb]{ 0,  .38,  0}{68.1 } & \cellcolor[rgb]{ .776,  .937,  .808}\textcolor[rgb]{ 0,  .38,  0}{40.9 } & \cellcolor[rgb]{ .776,  .937,  .808}\textcolor[rgb]{ 0,  .38,  0}{33.2 } \\
                  & RTX 3070 & \cellcolor[rgb]{ .776,  .937,  .808}\textcolor[rgb]{ 0,  .38,  0}{64.7 } & \cellcolor[rgb]{ .776,  .937,  .808}\textcolor[rgb]{ 0,  .38,  0}{64.6 } & \cellcolor[rgb]{ .776,  .937,  .808}\textcolor[rgb]{ 0,  .38,  0}{63.8 } & \cellcolor[rgb]{ .776,  .937,  .808}\textcolor[rgb]{ 0,  .38,  0}{61.0 } & \cellcolor[rgb]{ .776,  .937,  .808}\textcolor[rgb]{ 0,  .38,  0}{57.3 } & \cellcolor[rgb]{ .776,  .937,  .808}\textcolor[rgb]{ 0,  .38,  0}{52.6 } & \cellcolor[rgb]{ .776,  .937,  .808}\textcolor[rgb]{ 0,  .38,  0}{42.8 } & 26.1  & \cellcolor[rgb]{ 1,  .78,  .808}\textcolor[rgb]{ .612,  0,  .024}{OOM} \\
                  & RTX 3060 & \cellcolor[rgb]{ .776,  .937,  .808}\textcolor[rgb]{ 0,  .38,  0}{44.5 } & \cellcolor[rgb]{ .776,  .937,  .808}\textcolor[rgb]{ 0,  .38,  0}{44.3 } & \cellcolor[rgb]{ .776,  .937,  .808}\textcolor[rgb]{ 0,  .38,  0}{43.8 } & \cellcolor[rgb]{ .776,  .937,  .808}\textcolor[rgb]{ 0,  .38,  0}{42.1 } & \cellcolor[rgb]{ .776,  .937,  .808}\textcolor[rgb]{ 0,  .38,  0}{39.5 } & \cellcolor[rgb]{ .776,  .937,  .808}\textcolor[rgb]{ 0,  .38,  0}{36.3 } & 29.9  & 18.6  & 15.2  \\
                  & RTX 2080Ti & \cellcolor[rgb]{ .776,  .937,  .808}\textcolor[rgb]{ 0,  .38,  0}{93.1 } & \cellcolor[rgb]{ .776,  .937,  .808}\textcolor[rgb]{ 0,  .38,  0}{91.9 } & \cellcolor[rgb]{ .776,  .937,  .808}\textcolor[rgb]{ 0,  .38,  0}{89.9 } & \cellcolor[rgb]{ .776,  .937,  .808}\textcolor[rgb]{ 0,  .38,  0}{83.3 } & \cellcolor[rgb]{ .776,  .937,  .808}\textcolor[rgb]{ 0,  .38,  0}{74.7 } & \cellcolor[rgb]{ .776,  .937,  .808}\textcolor[rgb]{ 0,  .38,  0}{65.7 } & \cellcolor[rgb]{ .776,  .937,  .808}\textcolor[rgb]{ 0,  .38,  0}{49.8 } & 27.9  & 22.2  \\
                  & GTX 1080Ti & \cellcolor[rgb]{ .776,  .937,  .808}\textcolor[rgb]{ 0,  .38,  0}{50.0 } & \cellcolor[rgb]{ .776,  .937,  .808}\textcolor[rgb]{ 0,  .38,  0}{49.6 } & \cellcolor[rgb]{ .776,  .937,  .808}\textcolor[rgb]{ 0,  .38,  0}{48.8 } & \cellcolor[rgb]{ .776,  .937,  .808}\textcolor[rgb]{ 0,  .38,  0}{45.9 } & \cellcolor[rgb]{ .776,  .937,  .808}\textcolor[rgb]{ 0,  .38,  0}{41.5 } & \cellcolor[rgb]{ .776,  .937,  .808}\textcolor[rgb]{ 0,  .38,  0}{36.9 } & 28.2  & 15.9  & 12.6  \\
                  & GTX 1070Ti & 29.9  & 29.6  & 29.2  & 27.8  & 25.7  & 23.5  & 18.6  & 11.0  & \cellcolor[rgb]{ 1,  .78,  .808}\textcolor[rgb]{ .612,  0,  .024}{OOM} \\
            \hline
            \multirow{8}{*}{MCT-WCT$^2$} & A100-40G & 14.2  & 14.1  & 14.1  & 14.0  & 13.9  & 13.6  & 13.5  & 12.6  & 12.1  \\
                  & RTX 3090 & 13.4  & 13.3  & 13.2  & 12.9  & 12.9  & 12.7  & 12.5  & 12.3  & 11.6  \\
                  & RTX 3080 & 13.2  & 12.6  & 12.5  & 12.5  & 12.3  & 12.3  & 11.6  & 10.5  & 9.9  \\
                  & RTX 3070 & 11.9  & 11.9  & 11.8  & 11.5  & 11.5  & 11.0  & 10.9  & 9.4   & \cellcolor[rgb]{ 1,  .78,  .808}\textcolor[rgb]{ .612,  0,  .024}{OOM} \\
                  & RTX 3060 & 10.4  & 10.4  & 10.3  & 10.2  & 10.1  & 9.9   & 9.4   & 8.0   & 7.3  \\
                  & RTX 2080Ti & 14.4  & 14.4  & 14.4  & 14.2  & 14.1  & 13.7  & 12.9  & 10.6  & 9.8  \\
                  & GTX 1080Ti & 9.1   & 9.1   & 9.0   & 8.6   & 8.8   & 8.6   & 8.2   & 7.1   & 6.5  \\
                  & GTX 1070Ti & 7.5   & 7.4   & 7.4   & 7.4   & 7.2   & 7.1   & 6.6   & 5.4   & \cellcolor[rgb]{ 1,  .78,  .808}\textcolor[rgb]{ .612,  0,  .024}{OOM} \\
            \hline
            \multirow{8}{*}{MCT-GCANet} & A100-40G & \cellcolor[rgb]{ .776,  .937,  .808}\textcolor[rgb]{ 0,  .38,  0}{228.1 } & \cellcolor[rgb]{ .776,  .937,  .808}\textcolor[rgb]{ 0,  .38,  0}{226.1 } & \cellcolor[rgb]{ .776,  .937,  .808}\textcolor[rgb]{ 0,  .38,  0}{223.9 } & \cellcolor[rgb]{ .776,  .937,  .808}\textcolor[rgb]{ 0,  .38,  0}{221.1 } & \cellcolor[rgb]{ .776,  .937,  .808}\textcolor[rgb]{ 0,  .38,  0}{218.9 } & \cellcolor[rgb]{ .776,  .937,  .808}\textcolor[rgb]{ 0,  .38,  0}{197.4 } & \cellcolor[rgb]{ .776,  .937,  .808}\textcolor[rgb]{ 0,  .38,  0}{131.1 } & \cellcolor[rgb]{ .776,  .937,  .808}\textcolor[rgb]{ 0,  .38,  0}{61.3 } & \cellcolor[rgb]{ .776,  .937,  .808}\textcolor[rgb]{ 0,  .38,  0}{47.4 } \\
                  & RTX 3090 & \cellcolor[rgb]{ .776,  .937,  .808}\textcolor[rgb]{ 0,  .38,  0}{225.1 } & \cellcolor[rgb]{ .776,  .937,  .808}\textcolor[rgb]{ 0,  .38,  0}{222.1 } & \cellcolor[rgb]{ .776,  .937,  .808}\textcolor[rgb]{ 0,  .38,  0}{221.9 } & \cellcolor[rgb]{ .776,  .937,  .808}\textcolor[rgb]{ 0,  .38,  0}{213.0 } & \cellcolor[rgb]{ .776,  .937,  .808}\textcolor[rgb]{ 0,  .38,  0}{195.6 } & \cellcolor[rgb]{ .776,  .937,  .808}\textcolor[rgb]{ 0,  .38,  0}{165.8 } & \cellcolor[rgb]{ .776,  .937,  .808}\textcolor[rgb]{ 0,  .38,  0}{114.1 } & \cellcolor[rgb]{ .776,  .937,  .808}\textcolor[rgb]{ 0,  .38,  0}{56.1 } & \cellcolor[rgb]{ .776,  .937,  .808}\textcolor[rgb]{ 0,  .38,  0}{43.3 } \\
                  & RTX 3080 & \cellcolor[rgb]{ .776,  .937,  .808}\textcolor[rgb]{ 0,  .38,  0}{216.2 } & \cellcolor[rgb]{ .776,  .937,  .808}\textcolor[rgb]{ 0,  .38,  0}{215.4 } & \cellcolor[rgb]{ .776,  .937,  .808}\textcolor[rgb]{ 0,  .38,  0}{214.2 } & \cellcolor[rgb]{ .776,  .937,  .808}\textcolor[rgb]{ 0,  .38,  0}{194.9 } & \cellcolor[rgb]{ .776,  .937,  .808}\textcolor[rgb]{ 0,  .38,  0}{166.8 } & \cellcolor[rgb]{ .776,  .937,  .808}\textcolor[rgb]{ 0,  .38,  0}{139.5 } & \cellcolor[rgb]{ .776,  .937,  .808}\textcolor[rgb]{ 0,  .38,  0}{95.9 } & \cellcolor[rgb]{ .776,  .937,  .808}\textcolor[rgb]{ 0,  .38,  0}{46.3 } & \cellcolor[rgb]{ .776,  .937,  .808}\textcolor[rgb]{ 0,  .38,  0}{35.7 } \\
                  & RTX 3070 & \cellcolor[rgb]{ .776,  .937,  .808}\textcolor[rgb]{ 0,  .38,  0}{157.1 } & \cellcolor[rgb]{ .776,  .937,  .808}\textcolor[rgb]{ 0,  .38,  0}{155.4 } & \cellcolor[rgb]{ .776,  .937,  .808}\textcolor[rgb]{ 0,  .38,  0}{151.9 } & \cellcolor[rgb]{ .776,  .937,  .808}\textcolor[rgb]{ 0,  .38,  0}{134.5 } & \cellcolor[rgb]{ .776,  .937,  .808}\textcolor[rgb]{ 0,  .38,  0}{115.5 } & \cellcolor[rgb]{ .776,  .937,  .808}\textcolor[rgb]{ 0,  .38,  0}{97.1 } & \cellcolor[rgb]{ .776,  .937,  .808}\textcolor[rgb]{ 0,  .38,  0}{66.2 } & \cellcolor[rgb]{ .776,  .937,  .808}\textcolor[rgb]{ 0,  .38,  0}{32.0 } & \cellcolor[rgb]{ 1,  .78,  .808}\textcolor[rgb]{ .612,  0,  .024}{OOM} \\
                  & RTX 3060 & \cellcolor[rgb]{ .776,  .937,  .808}\textcolor[rgb]{ 0,  .38,  0}{110.7 } & \cellcolor[rgb]{ .776,  .937,  .808}\textcolor[rgb]{ 0,  .38,  0}{108.9 } & \cellcolor[rgb]{ .776,  .937,  .808}\textcolor[rgb]{ 0,  .38,  0}{106.4 } & \cellcolor[rgb]{ .776,  .937,  .808}\textcolor[rgb]{ 0,  .38,  0}{95.8 } & \cellcolor[rgb]{ .776,  .937,  .808}\textcolor[rgb]{ 0,  .38,  0}{82.6 } & \cellcolor[rgb]{ .776,  .937,  .808}\textcolor[rgb]{ 0,  .38,  0}{69.3 } & \cellcolor[rgb]{ .776,  .937,  .808}\textcolor[rgb]{ 0,  .38,  0}{48.1 } & 23.6  & 18.2  \\
                  & RTX 2080Ti & \cellcolor[rgb]{ .776,  .937,  .808}\textcolor[rgb]{ 0,  .38,  0}{194.0 } & \cellcolor[rgb]{ .776,  .937,  .808}\textcolor[rgb]{ 0,  .38,  0}{190.2 } & \cellcolor[rgb]{ .776,  .937,  .808}\textcolor[rgb]{ 0,  .38,  0}{186.5 } & \cellcolor[rgb]{ .776,  .937,  .808}\textcolor[rgb]{ 0,  .38,  0}{158.7 } & \cellcolor[rgb]{ .776,  .937,  .808}\textcolor[rgb]{ 0,  .38,  0}{131.1 } & \cellcolor[rgb]{ .776,  .937,  .808}\textcolor[rgb]{ 0,  .38,  0}{106.0 } & \cellcolor[rgb]{ .776,  .937,  .808}\textcolor[rgb]{ 0,  .38,  0}{70.0 } & \cellcolor[rgb]{ .776,  .937,  .808}\textcolor[rgb]{ 0,  .38,  0}{32.7 } & 24.8  \\
                  & GTX 1080Ti & \cellcolor[rgb]{ .776,  .937,  .808}\textcolor[rgb]{ 0,  .38,  0}{86.4 } & \cellcolor[rgb]{ .776,  .937,  .808}\textcolor[rgb]{ 0,  .38,  0}{84.9 } & \cellcolor[rgb]{ .776,  .937,  .808}\textcolor[rgb]{ 0,  .38,  0}{83.1 } & \cellcolor[rgb]{ .776,  .937,  .808}\textcolor[rgb]{ 0,  .38,  0}{73.8 } & \cellcolor[rgb]{ .776,  .937,  .808}\textcolor[rgb]{ 0,  .38,  0}{63.1 } & \cellcolor[rgb]{ .776,  .937,  .808}\textcolor[rgb]{ 0,  .38,  0}{52.5 } & \cellcolor[rgb]{ .776,  .937,  .808}\textcolor[rgb]{ 0,  .38,  0}{35.7 } & 17.5  & 13.4  \\
                  & GTX 1070Ti & \cellcolor[rgb]{ .776,  .937,  .808}\textcolor[rgb]{ 0,  .38,  0}{57.2 } & \cellcolor[rgb]{ .776,  .937,  .808}\textcolor[rgb]{ 0,  .38,  0}{56.2 } & \cellcolor[rgb]{ .776,  .937,  .808}\textcolor[rgb]{ 0,  .38,  0}{55.2 } & \cellcolor[rgb]{ .776,  .937,  .808}\textcolor[rgb]{ 0,  .38,  0}{49.7 } & \cellcolor[rgb]{ .776,  .937,  .808}\textcolor[rgb]{ 0,  .38,  0}{43.2 } & \cellcolor[rgb]{ .776,  .937,  .808}\textcolor[rgb]{ 0,  .38,  0}{36.7 } & 25.8  & 12.6  & \cellcolor[rgb]{ 1,  .78,  .808}\textcolor[rgb]{ .612,  0,  .024}{OOM} \\
            \hline
            \multirow{8}{*}{MCT-DPE} & A100-40G & \cellcolor[rgb]{ .776,  .937,  .808}\textcolor[rgb]{ 0,  .38,  0}{307.1 } & \cellcolor[rgb]{ .776,  .937,  .808}\textcolor[rgb]{ 0,  .38,  0}{301.3 } & \cellcolor[rgb]{ .776,  .937,  .808}\textcolor[rgb]{ 0,  .38,  0}{300.0 } & \cellcolor[rgb]{ .776,  .937,  .808}\textcolor[rgb]{ 0,  .38,  0}{297.3 } & \cellcolor[rgb]{ .776,  .937,  .808}\textcolor[rgb]{ 0,  .38,  0}{280.9 } & \cellcolor[rgb]{ .776,  .937,  .808}\textcolor[rgb]{ 0,  .38,  0}{252.1 } & \cellcolor[rgb]{ .776,  .937,  .808}\textcolor[rgb]{ 0,  .38,  0}{162.4 } & \cellcolor[rgb]{ .776,  .937,  .808}\textcolor[rgb]{ 0,  .38,  0}{72.9 } & \cellcolor[rgb]{ .776,  .937,  .808}\textcolor[rgb]{ 0,  .38,  0}{55.2 } \\
                  & RTX 3090 & \cellcolor[rgb]{ .776,  .937,  .808}\textcolor[rgb]{ 0,  .38,  0}{289.8 } & \cellcolor[rgb]{ .776,  .937,  .808}\textcolor[rgb]{ 0,  .38,  0}{288.9 } & \cellcolor[rgb]{ .776,  .937,  .808}\textcolor[rgb]{ 0,  .38,  0}{288.7 } & \cellcolor[rgb]{ .776,  .937,  .808}\textcolor[rgb]{ 0,  .38,  0}{288.3 } & \cellcolor[rgb]{ .776,  .937,  .808}\textcolor[rgb]{ 0,  .38,  0}{263.5 } & \cellcolor[rgb]{ .776,  .937,  .808}\textcolor[rgb]{ 0,  .38,  0}{224.4 } & \cellcolor[rgb]{ .776,  .937,  .808}\textcolor[rgb]{ 0,  .38,  0}{142.4 } & \cellcolor[rgb]{ .776,  .937,  .808}\textcolor[rgb]{ 0,  .38,  0}{63.7 } & \cellcolor[rgb]{ .776,  .937,  .808}\textcolor[rgb]{ 0,  .38,  0}{48.0 } \\
                  & RTX 3080 & \cellcolor[rgb]{ .776,  .937,  .808}\textcolor[rgb]{ 0,  .38,  0}{274.3 } & \cellcolor[rgb]{ .776,  .937,  .808}\textcolor[rgb]{ 0,  .38,  0}{273.4 } & \cellcolor[rgb]{ .776,  .937,  .808}\textcolor[rgb]{ 0,  .38,  0}{272.9 } & \cellcolor[rgb]{ .776,  .937,  .808}\textcolor[rgb]{ 0,  .38,  0}{265.4 } & \cellcolor[rgb]{ .776,  .937,  .808}\textcolor[rgb]{ 0,  .38,  0}{251.1 } & \cellcolor[rgb]{ .776,  .937,  .808}\textcolor[rgb]{ 0,  .38,  0}{198.6 } & \cellcolor[rgb]{ .776,  .937,  .808}\textcolor[rgb]{ 0,  .38,  0}{123.3 } & \cellcolor[rgb]{ .776,  .937,  .808}\textcolor[rgb]{ 0,  .38,  0}{53.8 } & \cellcolor[rgb]{ .776,  .937,  .808}\textcolor[rgb]{ 0,  .38,  0}{40.5 } \\
                  & RTX 3070 & \cellcolor[rgb]{ .776,  .937,  .808}\textcolor[rgb]{ 0,  .38,  0}{251.5 } & \cellcolor[rgb]{ .776,  .937,  .808}\textcolor[rgb]{ 0,  .38,  0}{251.4 } & \cellcolor[rgb]{ .776,  .937,  .808}\textcolor[rgb]{ 0,  .38,  0}{247.7 } & \cellcolor[rgb]{ .776,  .937,  .808}\textcolor[rgb]{ 0,  .38,  0}{226.6 } & \cellcolor[rgb]{ .776,  .937,  .808}\textcolor[rgb]{ 0,  .38,  0}{178.3 } & \cellcolor[rgb]{ .776,  .937,  .808}\textcolor[rgb]{ 0,  .38,  0}{138.4 } & \cellcolor[rgb]{ .776,  .937,  .808}\textcolor[rgb]{ 0,  .38,  0}{83.3 } & \cellcolor[rgb]{ .776,  .937,  .808}\textcolor[rgb]{ 0,  .38,  0}{35.6 } & \cellcolor[rgb]{ 1,  .78,  .808}\textcolor[rgb]{ .612,  0,  .024}{OOM} \\
                  & RTX 3060 & \cellcolor[rgb]{ .776,  .937,  .808}\textcolor[rgb]{ 0,  .38,  0}{208.5 } & \cellcolor[rgb]{ .776,  .937,  .808}\textcolor[rgb]{ 0,  .38,  0}{204.9 } & \cellcolor[rgb]{ .776,  .937,  .808}\textcolor[rgb]{ 0,  .38,  0}{196.0 } & \cellcolor[rgb]{ .776,  .937,  .808}\textcolor[rgb]{ 0,  .38,  0}{165.4 } & \cellcolor[rgb]{ .776,  .937,  .808}\textcolor[rgb]{ 0,  .38,  0}{130.0 } & \cellcolor[rgb]{ .776,  .937,  .808}\textcolor[rgb]{ 0,  .38,  0}{99.5 } & \cellcolor[rgb]{ .776,  .937,  .808}\textcolor[rgb]{ 0,  .38,  0}{60.4 } & 25.9  & 19.3  \\
                  & RTX 2080Ti & \cellcolor[rgb]{ .776,  .937,  .808}\textcolor[rgb]{ 0,  .38,  0}{274.1 } & \cellcolor[rgb]{ .776,  .937,  .808}\textcolor[rgb]{ 0,  .38,  0}{270.2 } & \cellcolor[rgb]{ .776,  .937,  .808}\textcolor[rgb]{ 0,  .38,  0}{267.7 } & \cellcolor[rgb]{ .776,  .937,  .808}\textcolor[rgb]{ 0,  .38,  0}{245.5 } & \cellcolor[rgb]{ .776,  .937,  .808}\textcolor[rgb]{ 0,  .38,  0}{185.6 } & \cellcolor[rgb]{ .776,  .937,  .808}\textcolor[rgb]{ 0,  .38,  0}{137.7 } & \cellcolor[rgb]{ .776,  .937,  .808}\textcolor[rgb]{ 0,  .38,  0}{80.8 } & \cellcolor[rgb]{ .776,  .937,  .808}\textcolor[rgb]{ 0,  .38,  0}{32.9 } & 24.8  \\
                  & GTX 1080Ti & \cellcolor[rgb]{ .776,  .937,  .808}\textcolor[rgb]{ 0,  .38,  0}{216.8 } & \cellcolor[rgb]{ .776,  .937,  .808}\textcolor[rgb]{ 0,  .38,  0}{211.5 } & \cellcolor[rgb]{ .776,  .937,  .808}\textcolor[rgb]{ 0,  .38,  0}{198.9 } & \cellcolor[rgb]{ .776,  .937,  .808}\textcolor[rgb]{ 0,  .38,  0}{152.6 } & \cellcolor[rgb]{ .776,  .937,  .808}\textcolor[rgb]{ 0,  .38,  0}{110.2 } & \cellcolor[rgb]{ .776,  .937,  .808}\textcolor[rgb]{ 0,  .38,  0}{80.3 } & \cellcolor[rgb]{ .776,  .937,  .808}\textcolor[rgb]{ 0,  .38,  0}{45.7 } & 18.5  & 14.0  \\
                  & GTX 1070Ti & \cellcolor[rgb]{ .776,  .937,  .808}\textcolor[rgb]{ 0,  .38,  0}{145.5 } & \cellcolor[rgb]{ .776,  .937,  .808}\textcolor[rgb]{ 0,  .38,  0}{141.4 } & \cellcolor[rgb]{ .776,  .937,  .808}\textcolor[rgb]{ 0,  .38,  0}{133.6 } & \cellcolor[rgb]{ .776,  .937,  .808}\textcolor[rgb]{ 0,  .38,  0}{105.6 } & \cellcolor[rgb]{ .776,  .937,  .808}\textcolor[rgb]{ 0,  .38,  0}{78.5 } & \cellcolor[rgb]{ .776,  .937,  .808}\textcolor[rgb]{ 0,  .38,  0}{58.3 } & \cellcolor[rgb]{ .776,  .937,  .808}\textcolor[rgb]{ 0,  .38,  0}{33.9 } & 13.8  & \cellcolor[rgb]{ 1,  .78,  .808}\textcolor[rgb]{ .612,  0,  .024}{OOM} \\
            \hline
            \end{tabular}%
    }
    \label{tab:speed}%
  \end{table*}%

\end{document}